\documentclass[10pt,twocolumn,letterpaper]{article}

\usepackage[pagenumbers]{cvpr} 

\definecolor{cvprblue}{rgb}{0.21,0.49,0.74}
\usepackage[pagebackref,breaklinks,colorlinks,allcolors=cvprblue]{hyperref}

\usepackage{colortbl}
\usepackage{graphicx}
\usepackage{diagbox}
\usepackage{multirow}
\usepackage{makecell}
\newcommand{\cb}{\cellcolor{blue!50}}
\newcommand{\clb}{\cellcolor{blue!20}}

\def\paperID{***}
\def\confName{CVPR}
\def\confYear{2026}

\title{Bringing Your Portrait to 3D Presence}

\author{
Jiawei Zhang\textsuperscript{\rm 1,2 $\dagger$}
\quad
Lei Chu\textsuperscript{\rm 2}
\quad
Jiahao Li\textsuperscript{\rm 2}
\quad
Zhenyu Zang\textsuperscript{\rm 2}
\quad
Chong Li\textsuperscript{\rm 2}
\\
Xiao Li\textsuperscript{\rm 2}
\quad
Xun Cao\textsuperscript{\rm 1}
\quad
Hao Zhu\textsuperscript{\rm 1}
\quad
Yan Lu\textsuperscript{\rm 2}
\vspace{5pt}\\
\textsuperscript{\rm 1}Nanjing University
\qquad
\textsuperscript{\rm 2}Microsoft Research Asia
}

\definecolor{myRed}{rgb}{0.18359375, 0.5625, 0.72265625}
\definecolor{myBlue}{rgb}{.0, .0, 1.0}

\begin{document}

\twocolumn[{%
\renewcommand\twocolumn[1][]{#1}%
\maketitle
\begin{center}
    \centering
\includegraphics[width=0.92\linewidth]{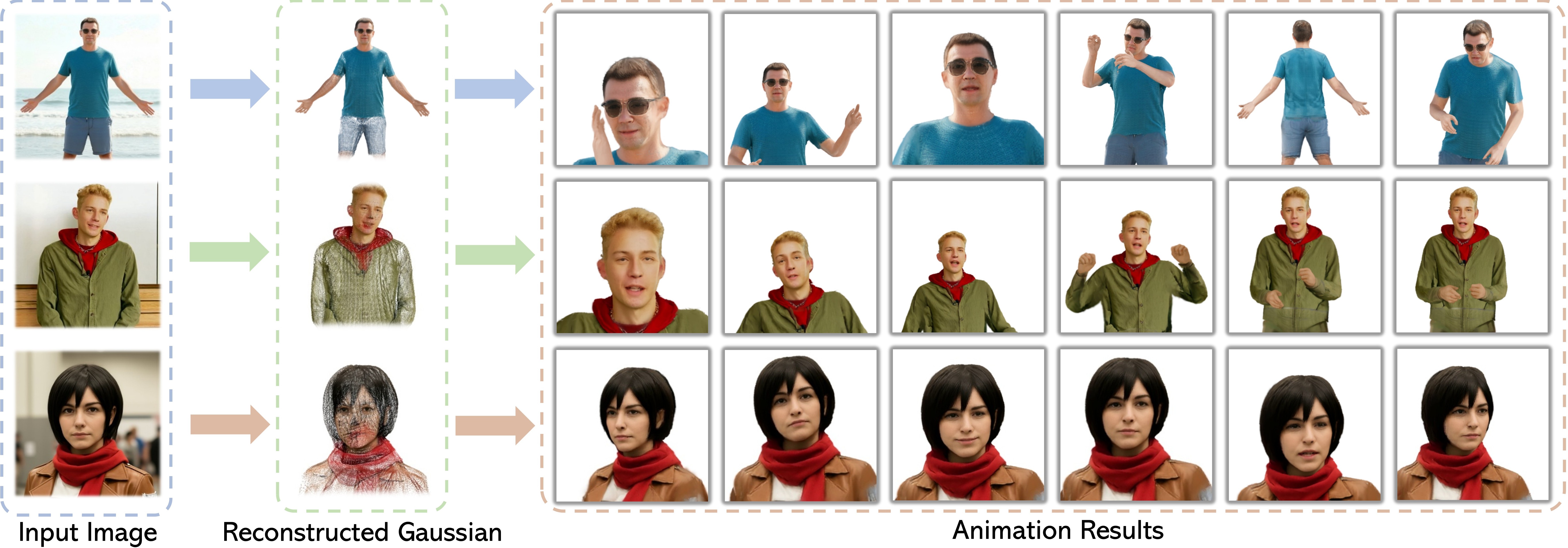}
    \captionof{figure}{Our method uses a dual-UV formulation to represent 3D avatars, enabling reconstruction from full-body, half-body, and headshot portraits while capturing off-body textures. Trained entirely on synthetic data, it generalizes effectively to in-the-wild images.}
\label{fig:teaser}
\end{center}
}]

\maketitle

\renewcommand{\thefootnote}{$\dagger$}
\footnotetext[1]{This work was done during Jiawei Zhang's internship at Microsoft Research Asia.}

\begin{abstract}
We present a unified framework for reconstructing animatable 3D human avatars from a single portrait across head, half-body, and full-body inputs. Our method tackles three bottlenecks: pose- and framing-sensitive feature representations, limited scalable data, and unreliable proxy-mesh estimation. We introduce a Dual-UV representation that maps image features to a canonical UV space via Core-UV and Shell-UV branches, eliminating pose- and framing-induced token shifts. We also build a factorized synthetic data manifold combining 2D generative diversity with geometry-consistent 3D renderings, supported by a training scheme that improves realism and identity consistency. A robust proxy-mesh tracker maintains stability under partial visibility. Together, these components enable strong in-the-wild generalization. Trained only on half-body synthetic data, our model achieves state-of-the-art head and upper-body reconstruction and competitive full-body results. Extensive experiments and analyses further validate the effectiveness of our approach.
\end{abstract}
    
\section{Introduction}
\label{sec:intro}
Creating \textit{animatable} 3D human avatars is central to telepresence and virtual reality. While high-quality avatars usually rely on multi-view capture or depth sensors, these setups limit scalability; reconstructing an animatable avatar from a \textit{single portrait} offers a far more accessible solution.

Despite rapid progress in 3D human reconstruction, most existing methods are designed either for head-only or for full-body avatars, and often rely on specific input assumptions. In particular, many pipelines assume full-body visibility (including both hands and feet) to obtain stable proxy mesh (\textit{e.g.}, SMPL-X~\cite{SMPL-X}, FLAME~\cite{FLAME}) fitting, an assumption that is not always satisfied in real scenarios, where upper-body or partial views are more typical. Recent transformer-based frameworks, including the Large Avatar Model~\cite{lam} (LAM) and Large Human Model~\cite{lhm} (LHM), follow the Large Reconstruction Model~\cite{lrm} (LRM) paradigm by encoding input images into patch-level features and using learnable tokens to query them through cross-attention. This design enables fast single-image reconstruction without explicit geometry or texture optimization but constrains the representation to the image feature space, making it difficult to generalize across incomplete inputs. As shown in LHM, performance degrades under pose variation or partial-body inputs, and even when trained on half-body data, the ambiguous definition of “half-body,” ranging from shoulder to waist or thigh crops, leads to inconsistent spatial correspondence and noticeable quality drops compared to full-body cases.
Our goal is to relax both input and data requirements, advancing toward in-the-wild 3D avatar reconstruction from everyday captures such as webcams or phone portraits, thereby extending the scalability of animatable avatar reconstruction beyond controlled environments.

Our investigation reveals that the fundamental obstacles to advancing single-image 3D avatar reconstruction arise primarily from three aspects. First, \textbf{representation design}. Most existing pipelines inherit ViT-based pretrained encoders, which lack strict translation invariance and thus require the input image to be spatially aligned to a fixed reference. Unlike general object reconstruction, human images exhibit large pose variations and frequent partial-body visibility, making such alignment inherently unstable.
Consequently, the decoder must learn to correlate image patches with 3D representations while adapting to token distribution shifts induced by pose and alignment inconsistencies, often leading to identity drift and texture distortion. Second, \textbf{data scalability}. High-quality multi-view human datasets require expensive studio setups with synchronized cameras, whereas real monocular videos demand intensive manual cleaning to ensure temporal and pose consistency. Synthetic data from traditional rendering engines offer controllable geometry but limited appearance diversity and a large domain gap to real imagery. Although 2D generative models can produce photorealistic humans with diverse appearances, their results generally lack identity and cross-view consistency, making them unsuitable for 3D supervision without further processing. Third, \textbf{robust body estimation}. Reliable proxy mesh tracking remains a key bottleneck. Existing trackers often assume full-body visibility, some even require both hands or the entire silhouette, to stabilize optimization, which rarely holds in in-the-wild captures dominated by upper-body views.

To address these challenges, we present a unified pipeline that integrates representation design, data construction, and proxy mesh estimation in a coherent framework.
(1) \textbf{Dual-UV Representation}.
At the core of our system is a Dual-UV representation that rearranges image features into a continuous, geometry-aligned UV space. It comprises two complementary branches: a Core-UV that encodes on-surface, geometry-anchored features, and a Shell-UV that captures off-surface details such as hair, clothing, and accessories by sampling features on an offset mesh shell. By anchoring tokens to a canonical surface rather than image coordinates, it eliminates token-distribution shifts caused by pose and alignment variations, which often lead to identity drift and texture distortion. This enables a single model to robustly handle head-only, half-body, and full-body inputs within one framework.
(2) \textbf{Factorized Synthetic Data Manifold}.
Our model is trained entirely on synthetic data that combines 2D generative and 3D rendered sources to achieve both appearance diversity and geometric reliability. Rather than enforcing multi-view consistency on 2D generative models, we leverage their ability to produce diverse, photorealistic appearances resembling real imagery. The 3D renderings, though less realistic, provide consistent geometric supervision that anchors reconstruction. All data are organized within a factorized, controllable manifold defined by semantically interpretable factors. A tailored training scheme mitigates identity and cross-view inconsistencies in the 2D data, while a realism regularizer projects each sample into a physically coherent, filmic space, preserving diversity while enhancing plausibility. Together, these designs yield a scalable synthetic corpus that enables stable training and strong generalization to in-the-wild captures.
(3) \textbf{Proxy Mesh Estimation}.
We empirically identify a stable configuration for human proxy mesh tracking under varying input completeness. Unlike previous approaches requiring full-body or both-hand visibility, our tracker maintains reliable performance across head-only, half-body, and full-body inputs, lowering capture demands and improving data scalability.

In summary, our work advances single-image 3D avatar reconstruction through the following contributions:
\begin{itemize}
    \item We propose a simple yet effective Dual-UV representation that maps all inputs into a continuous, geometry-aligned UV space, enabling a single model to handle head-only, half-body, and full-body views.
    \item We construct a factorized synthetic data manifold with rich appearance diversity and controllable structure, supported by a training scheme that reduces identity inconsistency and cross-view mismatch, enabling strong real-world generalization from synthetic data alone.
    \item We develop a robust proxy-mesh tracker that remains stable across varying input completeness, reducing dependency on full-body visibility and improving scalability for in-the-wild reconstruction.
    \item Trained solely on half-body data, our approach achieves state-of-the-art head and upper-body results and competitive full-body performance.
\end{itemize}

\begin{figure*}[]
    \centering
    \includegraphics[width=\textwidth]{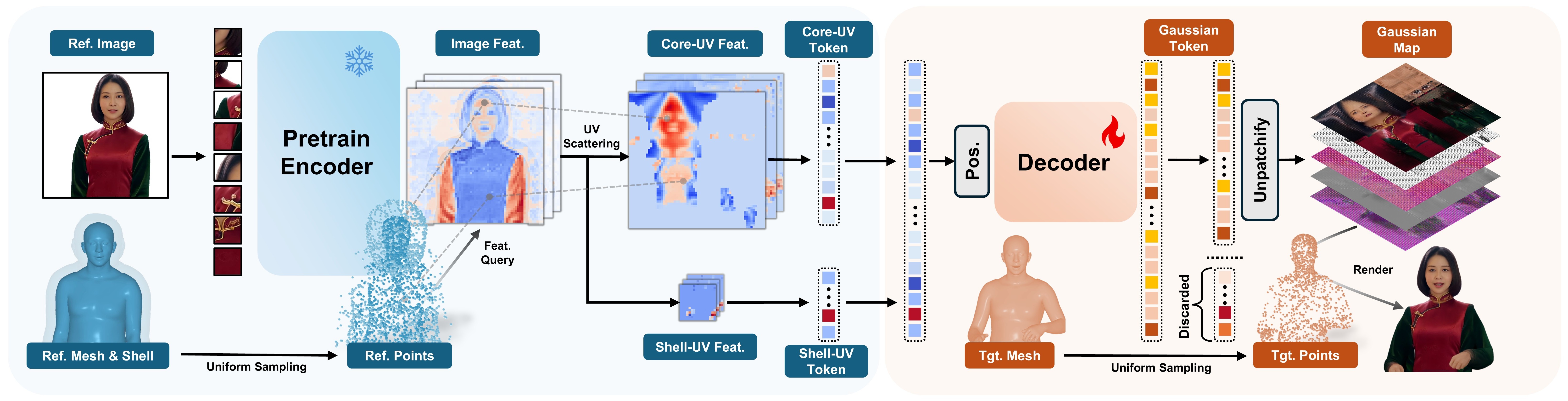}
    \vspace{-8mm}
    \caption{\textbf{Reconstruction Pipeline}.
    Given a reference image and its tracked proxy mesh, dense features from a frozen encoder are sampled along visible rays and scattered into canonical UV space to form the Core-UV map, while an offset shell captures off-surface regions such as hair and clothing. The Core-UV and Shell-UV tokens are fused and decoded by a lightweight transformer to reconstruct UV-space Gaussian attributes, which are then rigged to a target mesh and rendered from arbitrary viewpoints.
    }
    \vspace{-5mm}
    \label{fig: pipeline}
\end{figure*}

\section{Related Work}
\label{sec:relate}

\subsection{Human Datasets}
Large-scale datasets underpin modern human modeling and fall into two categories. Structured datasets~\cite{thuman,2k2k} use controlled multi-view capture and provide diverse expressions, poses, and lighting~\cite{wang2022faceverse,yang2020facescape,kirschstein2023nersemble,facewarehouse,actorhq,taoavatar,xiong2024mvhumannet,dnarendering,ava256,humanolat2025}. They support factor disentanglement but are costly and limited to studio environments. Unstructured datasets~\cite{ffhq,shhq,xie2022vfhq,tiktok,openhumanvid,tiktok2} collect in-the-wild images and videos with broad identity coverage but suffer from strong viewpoint bias toward near-frontal shots.

A complementary direction builds synthetic datasets~\cite{synbody,lookma,rodin} by rendering detailed assets with accurate annotations. They offer high controllability but exhibit realism and texture gaps relative to real imagery. Recent generative models~\cite{latentdiffusion,vqgan,idol,learn2control} help narrow this gap, enabling large-scale, realistic, and diverse synthetic human data. IDOL~\cite{idol} also leverages 2D generative models for data curation by finetuning them for multi-view synthesis, but this introduces bias to the data diversity and still lacks reliable geometric consistency. In contrast, we retain the raw texture richness and diversity of 2D generative data without relying on its multi-view validity.

\subsection{Human Modeling}
3D human modeling spans head and full-body reconstruction, following a similar evolution. Early methods recovered geometry with explicit meshes or implicit fields~\cite{xiu2023econ,pifu,sifu,xiu2022icon}, and neural rendering~\cite{mildenhall2020nerf,kerbl3Dgaussians} shifted the focus to appearance. 3D-aware GANs~\cite{piGAN2021,eg3d} enabled controllable head~\cite{spherehead,panohead,li2025hyplanehead,gghead,yu25gaia,realityavatar} and body~\cite{zhang2023getavatar,EVA3D,dong2023ag3d,gsm} synthesis, often leveraging 3DMM priors~\cite{3dmm}. Score-distillation~\cite{poole2022dreamfusion} further allowed diffusion-driven generation of diverse 3D humans~\cite{tang2023dreamgaussian,kolotouros2023dreamhuman,zhou2024headstudio,liu2024humangaussian,expressivetalkingbody,tera}.

A complementary line learns person-specific avatars from studio captures~\cite{debevec2012light} or monocular videos~\cite{firstmono}. For heads, studio setups enable generalizable models~\cite{li2024uravatar,sega,headnerf, wu2026uikafastuniversalhead}, while single-view methods have advanced controllable 3D head synthesis~\cite{liao2025soap, deng2024portrait4d, deng2024portrait4dv2, tran2024voodoo, zhang2024rodinhd, rome, joker2025, shi2025dex}  and per-subject optimization achieves high fidelity~\cite{imavatar,pointavatar,gaussianheadavatar,qian2023gaussianavatars,saito2024rgca,hybridgsavatar,expressivetalkingbody,flashavatar,zhang2025fate}. For bodies, appearance and clothing variability make generalization harder, so many works rely on per-subject fitting from monocular~\cite{vid2avatar,moon2024exavatar,neuralbody} or multi-view captures~\cite{relightablefullbodyavatarzju,relightfullbodymeta,animatablegaussian,dreamliftanimate,pan2024humansplat,sketch2pose}.

More recent efforts use multi-view~\cite{li2024pshuman,Kant2025Pippo,adahuman,chen2025synchuman} and video-diffusion models~\cite{qiu2024AniGS,lu2025gas,diffuman4d} to generate multi-view imagery and reconstruct 3D avatars. In parallel, methods~\cite{lhm,lam} based on large reconstruction models~\cite{lrm,pf_lrm} provide single-forward 3D inference.

Recent works \cite{gsm,gghead,li2024uravatar,idol} also use UV space for avatar attribute, but rely on StyleGAN or cross-attention to modulate UV attributes. We instead employ a non-learnable, closed-form projection from image space to UV space, enabling faithful texture transfer and strong identity preservation without generative hallucination.

Upper- or half-body avatar modeling from casual inputs is relatively underexplored and remains challenging. Existing half-body approaches are limited: many depend on fixed multi-view capture for real-time systems~\cite{telealoha,starline}, while others only extend head models slightly toward the shoulders~\cite{Portrait3D_sig24,3dportraitgan,animportrait3d,yue2023aniportraitgan,portraitgen}. Diffusion-based methods~\cite{lin2025cyberhost,guan2024talkact,liu2025tango} generate person-specific upper-body avatars but are generally constrained to frontal poses and do not generalize well to wider viewpoints. Recently, GUAVA~\cite{guava} achieved a generalizable upper-body Gaussian avatar. GUAVA also uses a UV branch for appearance, but unlike our dual-UV design, it relies on a separate template branch to encode the input, and stabilize the training process, thus requires a later refining module to blend the Gaussians from two branches.
\section{Method}
\label{sec:method}
\begin{figure*}[]
    \centering
    \includegraphics[width=\textwidth]{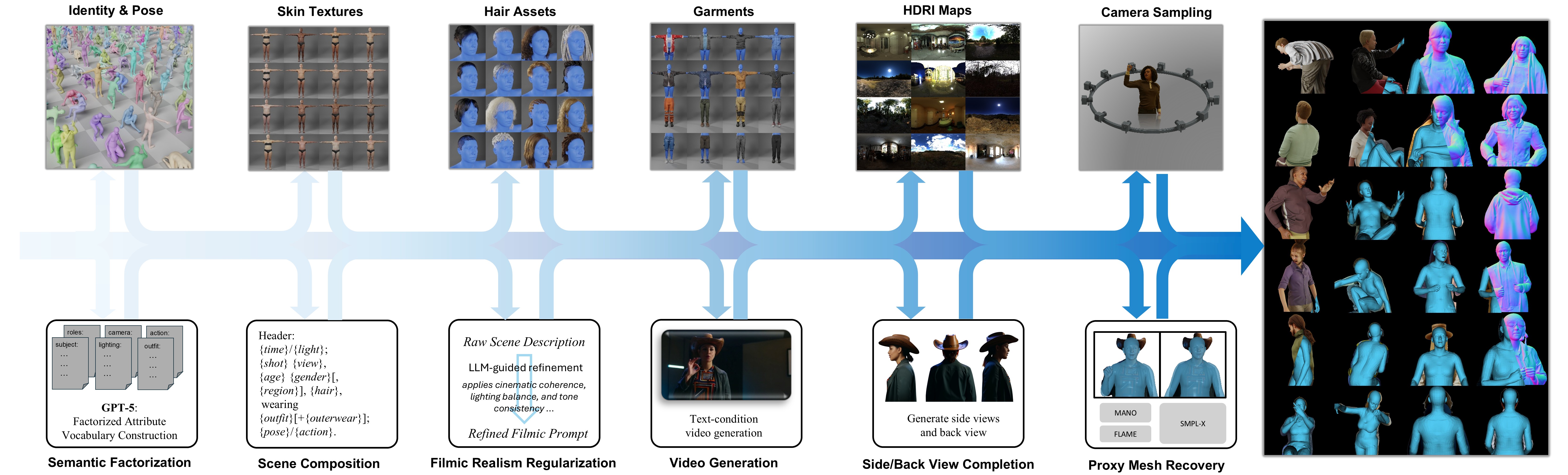}
    \vspace{-7mm}
    \caption{\textbf{Data Curation}.
    We build a hybrid dataset by combining geometry-anchored 3D rendering with semantics-driven generative synthesis.
The \textit{synthetic rendering branch} offers geometry-consistent multi-view supervision through procedural sampling of identity, pose, appearance, illumination, and cameras.
The \textit{generative branch} constructs a factorized appearance manifold by decomposing scene attributes, applying LLM-based filmic refinement, generating photorealistic sequences, and completing each sample with side/back views for weakly correlated augmentation.
}
 \vspace{-7mm}
    \label{fig:data}
\end{figure*}

Given a single RGB portrait  $I$, our goal is to reconstruct an \textit{animatable} 3D human avatar represented by a set of Gaussians
$
\mathcal{G} = \{ g_i = (\boldsymbol{\mu}_i, \Sigma_i, \mathbf{c}_i, \alpha_i) \}_{i=1}^{N}.
$
We decompose the latent space into two complementary parts: $\mathbf{z} = \{ \mathbf{z}_{\text{uv}}, \mathbf{z}_{\text{mesh}} \},$ where $\mathbf{z}_{\text{uv}}$ is a \textit{dual-UV representation} encoding geometry-aligned appearance and visibility in a canonical UV space, and $\mathbf{z}_{\text{mesh}}$ is a \textit{proxy mesh latent} parameterized by proxy mesh to capture pose-dependent deformation. 
The overall mapping is formulated as
\[
f_\theta: I \rightarrow \mathbf{z}_{\text{uv}}, 
\quad
\mathcal{G} = \Phi(\mathbf{z}_{\text{uv}}, \mathbf{z}_{\text{mesh}}),
\]
where $f_\theta$ denotes the reconstruction network and $\Phi$ converts latent codes into posed 3D Gaussians.

In Sec.~\ref{sec:reconstruction_model}, we present the mask-based reconstruction model, centered on the dual-UV representation for robust alignment across varying input completeness, as shown in Fig.~\ref{fig: pipeline}. 
Sec.~\ref{sec:data_curation} details the factorized synthetic data manifold curation pipeline that provides scalable and diverse supervision.
Finally, Sec.~\ref{sec:mesh} introduces the proxy mesh estimation framework for stable proxy mesh tracking. Refer to the supplementary for implementation and loss details.

\subsection{Mask-based Reconstruction}
\label{sec:reconstruction_model}

Large reconstruction models~\cite{lrm,pf_lrm} map images to 3D by querying patch features with learnable tokens, which works for generic objects but entangles pose and identity for animatable humans. Because inputs vary greatly, from head-only to full-body, token correspondences become sensitive to framing and alignment, harming generalization. Moreover, these models must implicitly recover the canonical structure from posed images, even though the proxy mesh already provides the deformation field. This forces the decoder to relearn geometric mappings and wastes capacity.

We instead reconstruct through a geometry-aligned \textit{Dual-UV representation} that deterministically maps image features into canonical UV space. This removes pose ambiguity, and lets the network focus solely on identity and appearance detail. The Dual-UV representation contains two complementary components.

\paragraph{Core-UV feature encoding.}
The Core-UV branch establishes a deterministic correspondence between image pixels and the canonical mesh surface. 
Using the UV layout of the human mesh, we define a differentiable unparameterization
\begin{equation}
(u,v)=\mathcal{M}^{-1}(\mathbf{p};M),
\end{equation}
where each surface point $\mathbf{p}$ on mesh $M$ is uniquely mapped to its UV coordinate $(u,v)$.
We uniformly sample surface points $\{\mathbf{p}_i\}_{i=1}^{N}$ on $M$ and precompute their face indices and barycentric coordinates. 

Given an input image $I$ and calibrated camera $\Pi$, we follow previous works~\cite{lhm,idol}, use a frozen Sapiens-1B encoder~\cite{khirodkar2024sapiens} to extract image features 
$\mathbf{F}=\mathcal{E}(I)$.
We rasterize $M$ under $\Pi$ with back-face culling and z-buffering to determine visibility.
For each visible point $\mathbf{p}_i$, it is projected to the image $\mathbf{x}_i=\Pi(\mathbf{p}_i)$,
and its feature is gathered by:
\begin{equation}
\mathbf{f}_i=\mathcal{S}(\mathbf{F},\mathbf{x}_i),\quad m_i\in\{0,1\},
\end{equation}
where 
$\mathcal{S}$ denotes differentiable bilinear sampling that interpolates local features from 
$\mathbf{F}$ at subpixel coordinate $\mathbf{x}_i$. The sampled features are then scattered onto a regular UV grid 
$\tilde{\mathbf{U}}\in\mathbb{R}^{H_U\times W_U\times C}$:
\begin{equation}
\tilde{\mathbf{U}}(u,v)
=\frac{\sum_i m_i\,k((u,v)-(u_i,v_i))\,\mathbf{f}_i}
{\sum_i m_i\,k((u,v)-(u_i,v_i))+\varepsilon},
\end{equation}
where $k(\cdot)$ is a compact aggregation kernel (nearest-neighbor in practice) and $\varepsilon$ ensures stability.
This produces a canonical, geometry-aligned Core-UV feature map that provides a one-to-one, differentiable link between image observations and mesh-surface coordinates.

\paragraph{Shell-UV feature encoding.}
The human mesh $M$ captures only the body surface and cannot represent volumetric details such as hair or loose garments. 
To encode these off-surface regions, we construct an auxiliary shell $M^{+}$ by offsetting each mesh vertex along its outward normal:
$
M^{+} = \{\mathbf{p} + \delta\,\mathbf{n}(\mathbf{p}) \mid \mathbf{p}\in M\},
$
where $\mathbf{n}(\mathbf{p})$ is the vertex normal and $\delta$ is a small offset.
We rasterize both $M$ and $M^{+}$ under camera $\Pi$ to obtain their visibility masks 
$m_M$ and $m_{M^{+}}$.
The \textit{shell-only} region is defined as the visible area of $S$ excluding the projection of $M$,
$
m_{\text{shell}} = m_{M^{+}} \cdot (1 - m_M),
$
ensuring that only off-surface pixels are used for feature sampling. For $m_{\text{shell}}$, it is multiplied by the body mask rendered from the shell proxy mesh.

For each visible point $\mathbf{p}_j\!\in\!M^{+}$ with $m_{\text{shell},j}=1$, 
its projection $\mathbf{x}_j=\Pi(\mathbf{p}_j)$ queries the image feature map $\mathbf{F}$ by differentiable bilinear sampling:
\begin{equation}
\mathbf{f}_j = \mathcal{S}(\mathbf{F}, \mathbf{x}_j).
\end{equation}
The sampled features are transferred to their corresponding UV coordinates 
$(u_j,v_j)=\mathcal{M}^{-1}(\mathbf{p}_j)$ and aggregated on a coarse UV grid 
$\tilde{\mathbf{U}}_{\text{shell}}\!\in\!\mathbb{R}^{H'_U\times W'_U\times C}$ using the same kernel aggregation as in the Core-UV branch.  
Although the mapping from $M^{+}$ to UV space is approximate, it maintains local spatial coherence and provides a soft positional prior for encoding off-surface appearance.

\paragraph{Decoder Design.}
Core-UV and Shell-UV tokens are concatenated, processed by a shallow transformer stack, and unpatched to form the UV attributes. Separate heads predict Gaussian attributes in UV space, including color $\boldsymbol{c}$, opacity $o$, offset $\boldsymbol{d}$, and rotation $\boldsymbol{r}$ and scale $\boldsymbol{s}$.

\paragraph{Discussion.}
Transformer-based avatar models~\cite{lhm,idol} typically employ large decoders with learnable queries, treating the pretrained encoder as a passive feature extractor and discarding the masked-autoencoding training asymmetry, where a strong encoder enables a lightweight decoder, as in Sapiens~\cite{khirodkar2024sapiens}. In contrast, we retain this asymmetry: projected image features fill only visible UV cells, leaving occluded regions blank, analogous to masked tokens. A compact transformer propagates information from visible to missing areas. This MAE-aligned design exploits pretrained reconstruction priors in geometry-aligned space, achieving high-quality avatars with far fewer decoder parameters than query-based approaches.

\subsection{Dataset Curation}
\label{sec:data_curation}
Training a single-image avatar reconstructor requires large-scale data that jointly cover geometric reliability and photorealistic diversity. 
Existing multi-view human datasets~\cite{actorhq,ava256,Kant2025Pippo} are limited in identity and appearance due to costly studio capture.
We therefore synthesize data from two complementary sources: a geometry-accurate \textit{synthetic rendering branch} and a photorealistic \textit{generative branch} (Fig.~\ref{fig:data}).

\noindent\textbf{Synthetic Rendering Branch.}
We render multi-view human images with a parametric body model following~\cite{lookma}. 
Identity, pose, garment, and lighting are sampled procedurally, and each subject is rendered from multiple calibrated viewpoints under HDRI environments.
This provides geometry-consistent supervision without manual annotation, forming the structural backbone of training data.

\noindent\textbf{Generative Branch.}
While prior efforts~\cite{learn2control,idol} fine-tune diffusion models for view-consistent humans, such constraints often degrade realism and diversity. 
We instead embrace a different philosophy: rather than forcing 2D generators to be multi-view consistent, we exploit their strengths, rich identity variation, natural appearance, and realistic motion, to populate a broad and controllable distribution.

To this end, we define a \textit{factorized data manifold} in which each sample is described by interpretable dimensions such as time of day, lighting, shot size, composition, clothing, hairstyle, role, region, and action.
Combinations of these factors are first assembled into concise textual descriptions by GPT-5~\cite{openai_gpt5_2025}, then refined by a large language model (Qwen2.5-14B-Instruct~\cite{qwen25}) acting as a \textit{realism regularizer}.
This refinement step projects each prompt into a physically coherent, filmic space—resolving contradictory attributes and enriching it with cinematographic cues on framing, illumination, and tone.
The refined prompts are passed to the text-to-video generator Wan2.2~\cite{wan22} to produce short, temporally consistent human clips.
From each clip, one representative frame is selected and complemented by side and back views synthesized through Qwen-Image-Edit~\cite{qwenimage}. Unlike prior works that assume perfect multi-view consistency, we treat these generated frames as weakly correlated views rather than ground-truth correspondences. During training, we impose a \textit{directional cross-view consistency} that flows only from more reliable to less reliable views (e.g., side$\!\rightarrow\!$back, front$\!\rightarrow\!$back), avoiding cyclic constraints that can amplify identity drift or texture aliasing. 
This asymmetric design effectively stabilizes training while still encouraging view-aware coherence.

\noindent\textbf{Discussion.}
Our design deliberately avoids fine-tuning generative models for 3D consistency, focusing instead on realism and diversity as complementary to the geometry-rich synthetic renders.
Together, these two branches form a scalable corpus where geometry and photorealism are disentangled yet coherent:
(i) the synthetic branch anchors geometric supervision,
(ii) the generative branch expands appearance coverage within a factorized manifold,
and (iii) the realism regularizer maintains filmic plausibility.
This combination enables robust training and strong generalization to in-the-wild portraits. \textit{Implementation details of the control factors, prompt templates, and LLM refinement commands are provided in the supplementary material.}

\subsection{Proxy Mesh Estimation}
\label{sec:mesh}

A reliable proxy mesh is essential for canonical avatar reconstruction, yet many pretrained models can provides initial estimates, but each of them assumes its own cropped views and coordinate system. We therefore analyze the stability ranges of multiple estimators and build a unified tracking pipeline rather than relying on a single model.

We benchmark representative estimators across head-only, half-body, and full-body inputs. OSX performs well with half-body visibility; Multi-HMR delivers accurate estimates when the entire body is visible; EMICA remains stable for head-dominant inputs; HaMeR provides accurate hand articulation only when hands are visible. This yields an empirical map of each model’s reliable operating regime. Guided by this map, we design a hierarchical framework that activates and fuses outputs from multiple estimators according to detected visibility, then jointly refine via keypoint reprojection and dense vertex alignment.

This tracker produces stable, anatomically coherent meshes for both real and generated images across all input types, from head-only portraits to full-body captures, and serves as a robust foundation for our reconstruction pipeline. Implementation details and further analyses are provided in the supplementary material.

\section{Experiment}
\label{sec:exp}

\begin{figure*}[!htbp]
    \centering
    \includegraphics[width=0.9\textwidth]{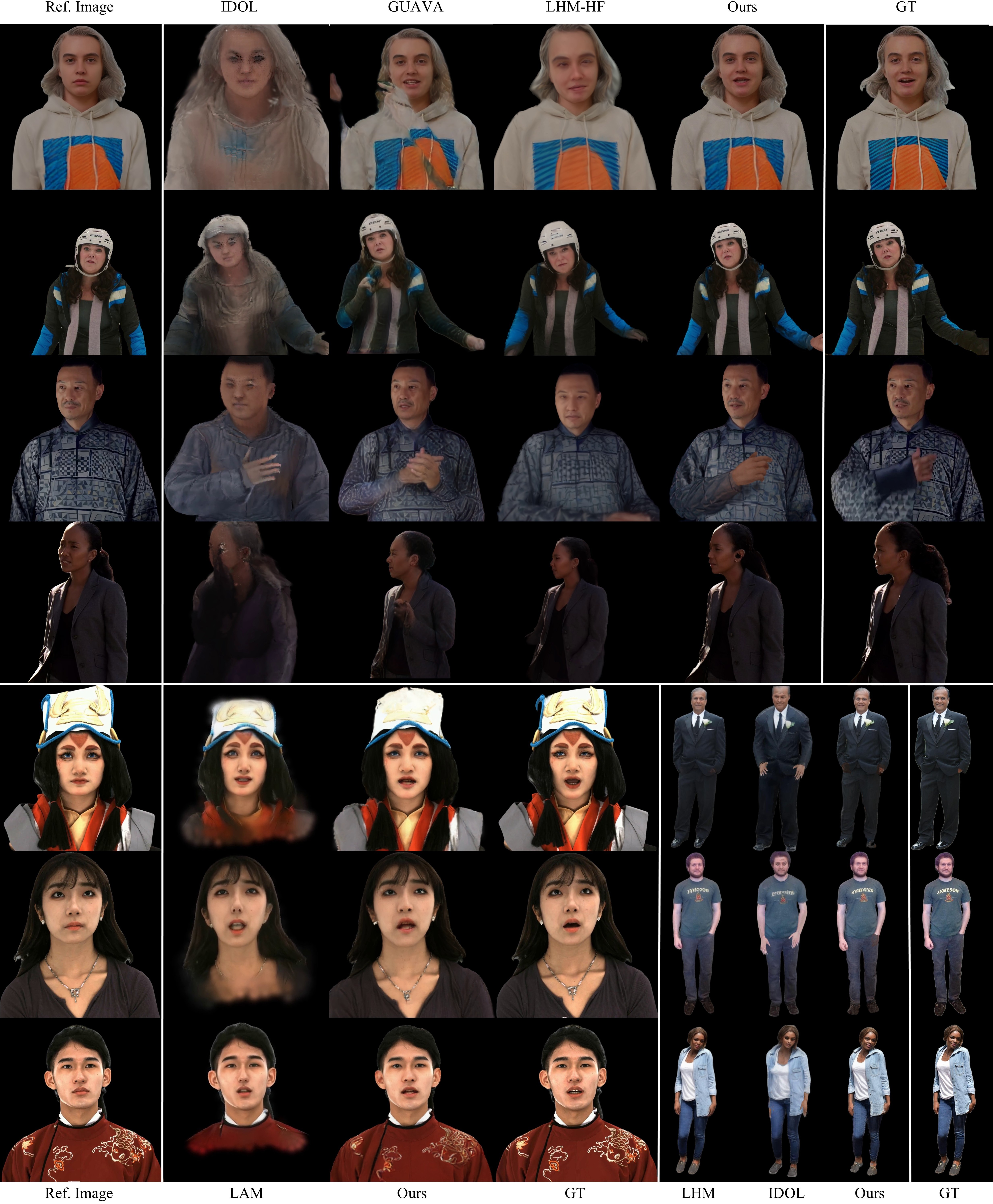}
    \vspace{-1em}
    \caption{\textbf{Reenactment Results}.
    Our method is trained solely on \textit{upper-body data}, generalizes well to head and full-body inputs.
    }
    \label{fig: self_reenact}
    \vspace{-8px}
\end{figure*}

\begin{figure*}[!htbp]
    \centering
    \includegraphics[width=0.95\textwidth]{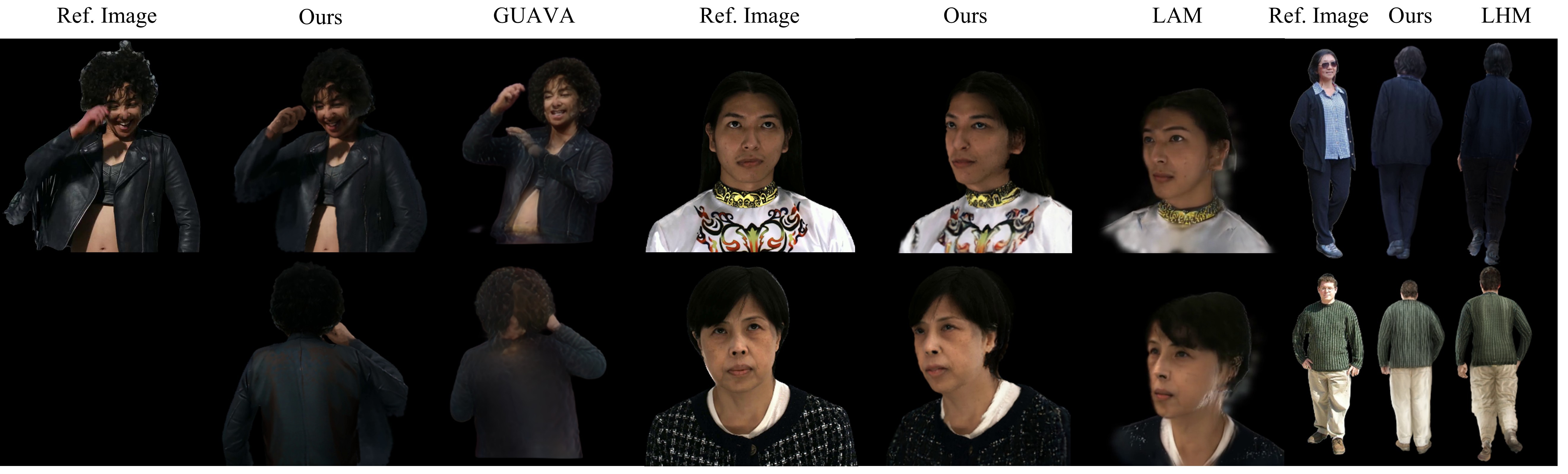}
    \vspace{-3mm}
    \caption{\textbf{Novel View Synthesis}.
    Our method generates multi-view human renderings from a single reference image, showing comparatively more consistent appearance, especially in the head and upper-body regions.
    }
    \label{fig: novel_view}
    \vspace{-3mm}
\end{figure*}

\begin{table*}[!htbp]

\caption{
\centering
Comparision of quantitative results with state-of-the-art methods.
}
\label{tab: self_reenact}

\vspace{-0.1in}
\centering
\small
\scalebox{0.83}{

\begin{tabular}{l|lll|lll|l|lll|l|lll}
\hline
\multirow{2}{*}{\diagbox[width=1.6cm, height=0.7cm]{}{}} & \multicolumn{3}{c|}{Upper-Wild} & \multicolumn{3}{c|}{Upper-Wan} & \multirow{2}{*}{\diagbox[width=1.2cm, height=0.7cm]{}{}}      & \multicolumn{3}{c|}{Full-Body} & \multirow{2}{*}{\diagbox[width=1.2cm, height=0.7cm]{}{}}      & \multicolumn{3}{c}{Head} \\ \cline{2-7} \cline{9-11} \cline{13-15} 
                  & PSNR↑    & SSIM↑     & LPIPS↓   & PSNR↑    & SSIM↑    & LPIPS↓   &                        & PSNR↑    & SSIM↑    & LPIPS↓   &                        & PSNR↑  & SSIM↑  & LPIPS↓ \\ \hline
Ours              & \cb{21.09}    & \cb{0.8510}    & \cb{0.1426}   & \cb{20.38}    & \cb{0.7867}   & \cb{0.1635}   & Ours                   & \cb{24.53}    & 0.8642   & \clb{0.0916}   & Ours                   & \cb{19.04}  & \cb{0.8526} & \cb{0.1613} \\
GUAVA             & \clb{20.59}    & \clb{0.7864}    & \clb{0.1957}   & \clb{20.24}    & \clb{0.7215}   & \clb{0.1940}   & IDOL                   & 18.51    & \clb{0.8753}   & 0.1256   & LAM$^\dagger$          & \clb{17.19}  & \clb{0.7526} & \clb{0.2207} \\
LHM-HF            & 13.95    & 0.7835    & 0.3335   & 12.04    & 0.6664   & 0.3626   & LHM                    & \clb{21.53}    & \cb{0.9151}   & \cb{0.0725}   & LAM                    & 14.81  & 0.6789 & 0.2613 \\
IDOL              & 12.93    & 0.7617    & 0.3465   & 10.35    & 0.5802   & 0.5132   & \multicolumn{1}{c|}{------} & \multicolumn{3}{c|}{----------------------------}         & \multicolumn{1}{c|}{------} & \multicolumn{3}{c}{----------------------------}    \\ \hline
\end{tabular}

}

\noindent
Note: \colorbox{blue!50}{blue} and \colorbox{blue!20}{lightblue} indicate the best and second-best results.
$\dagger$ indicates non-facial parts are parsed out.

\end{table*}
\subsection{Implementation Details}
We train on an upper-body–dominant synthetic dataset, while our data generation and training pipeline also flexibly accommodate full-body samples and head-only data. Our rendering pipeline generates 150K subjects assembled from curated 3D assets, each rendered from 12 upper-body–focused views. An additional 300K portrait video clips with talking or upper-body motion are synthesized, along with side and back views for a random frame. Data are split 19:1 for training and validation.

Our decoder has 8 self-attention layers and fewer than 0.1B parameters. Its outputs are unpatchified into $8\times8$ patches to form $512\times512$ Gaussian attribute maps (262K Gaussians). We train with AdamW~\cite{adam} at a learning rate of $1\times10^{-4}$, using mixed precision and gradient clipping (norm 1.0). Training runs for three days on 4 NVIDIA A100~80G GPUs with a batch size of 8 per GPU, the overall training cost is \textit{significantly lower} than LRM-based methods.

\subsection{Experiment Setup}
We compare with LHM, LHM-HF, IDOL, LAM, and GUAVA under head, upper-body, and full-body inputs.

LHM targets full-body reconstruction, encoding facial and body regions separately using DINOv2~\cite{oquab2023dinov2} and Sapiens backbones for cross-attention between modalities.
LHM-HF extends this design by training on half-body data augmented by random cropping from LHM’s large-scale in-the-wild video dataset.
IDOL applies a large Transformer decoder along with a heavy CNN-based decoder to regress Gaussian attributes from learnable tokens.
LAM simplifies LHM’s architecture, focusing on head modeling using both studio-captured~\cite{kirschstein2023nersemble} and in-the-wild data~\cite{xie2022vfhq}.
GUAVA specializes in upper-body reconstruction, featuring a dual-branch Gaussian decoder followed by a screen-space CNN refinement~\cite{styleavatar} stage for improved fidelity.


For upper-body evaluation, we use 100 real clips from OpenHumanVid~\cite{openhumanvid} and 200 synthesized clips; for head portraits, we sample 50 talking clips from RenderMe360~\cite{renderme360}; for full-body, we select 100 subjects from SHHQ~\cite{shhq}. We compare against the relevant baselines in each setting, using their own preprocessing pipeline.

\subsection{Comparison Results}
Quantitative results are reported in Tab.~\ref{tab: self_reenact} and qualitative comparisons in Fig.~\ref{fig: self_reenact}. For upper-body, our method surpasses LRM-based approaches (IDOL, LHM-HF) in texture fidelity and identity preservation.
While GUAVA is competitive on visible regions, its strict requirement for visible hands causes unstable poses and artifacts under hand occlusions. In contrast, our method avoids any 2D refinement and remains robust across varying visibility conditions.

For head reconstruction, we outperform the head-specific LAM in identity preservation and additionally reconstruct regions below the shoulders that LAM cannot represent.
In the full-body setting, our approach achieves performance comparable to dedicated single-image full-body methods, despite being trained almost exclusively on upper-body data and never seeing lower-body regions.
This indicates strong generalization to unseen body parts.

As shown in Fig.~\ref{fig: novel_view}, across all three settings our method produces novel views with consistent geometry and texture under diverse viewpoints and challenging poses.

\begin{figure}[!htbp]
    \centering
    \includegraphics[width=0.5\textwidth]{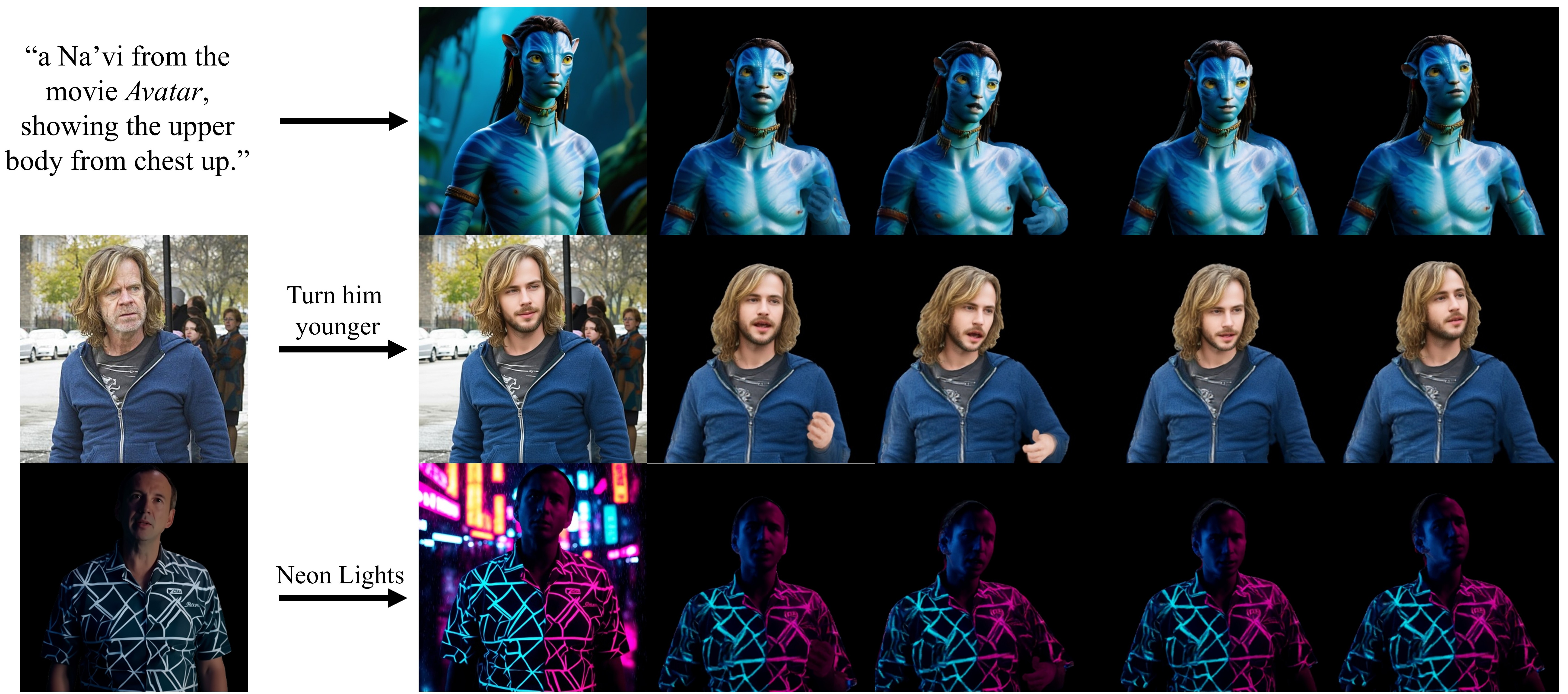}
    \vspace{-5mm}
    \caption{\textbf{Editing Results}.
    Our model supports various appearance edits from a single image,
    demonstrating its adaptability to diverse visual conditions.
    }
    \label{fig: edit}
    \vspace{-5mm}
\end{figure}

\begin{figure}[!htbp]
    \centering
    \includegraphics[width=0.5\textwidth]{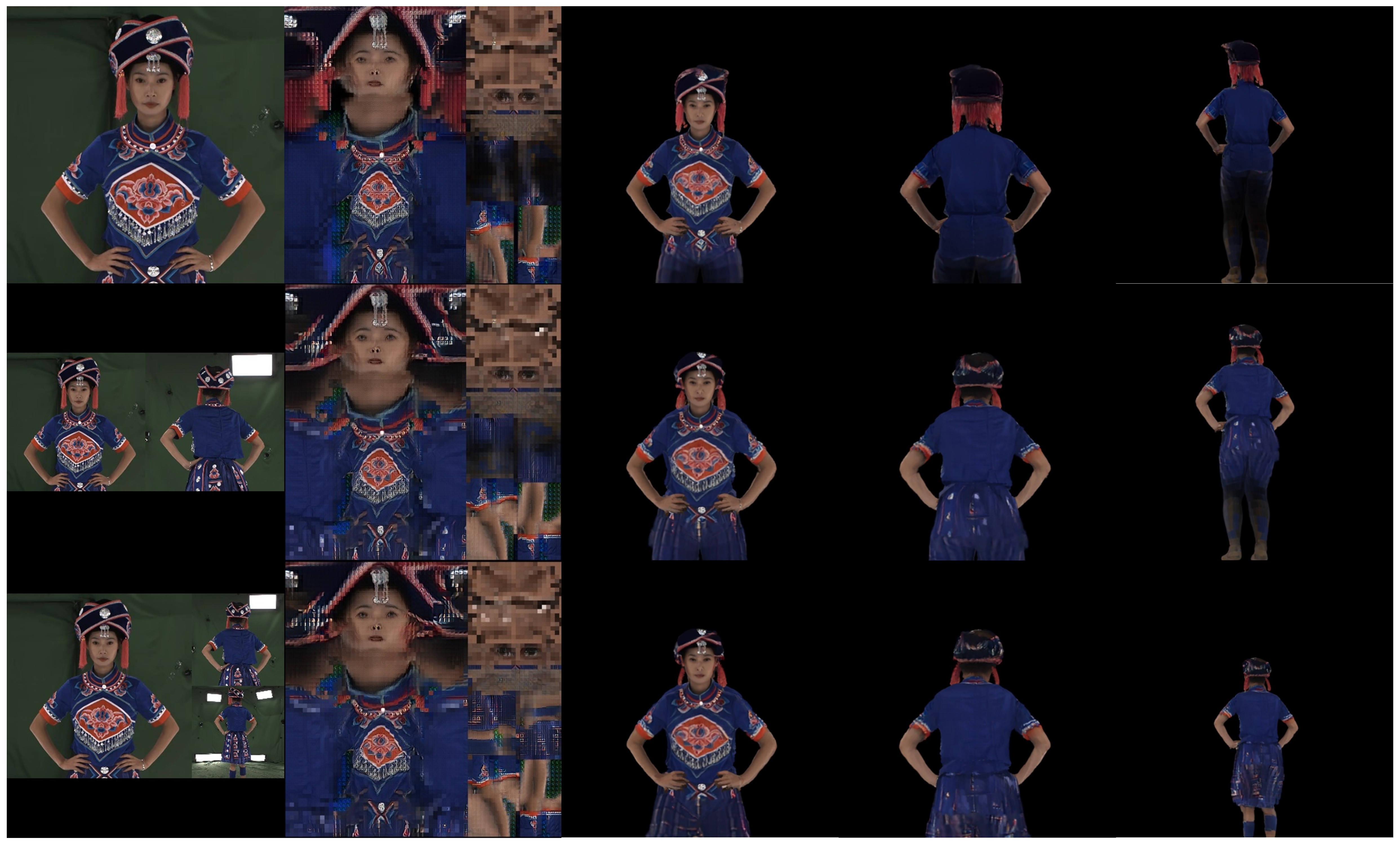}
    \vspace{-5mm}
    \caption{\textbf{Multiple Input}.
    Our model is capable of taking multiple images as input,
    indicating its potential flexibility in leveraging multi-view information.
    }
    \label{fig: multi}
    \vspace{-5mm}
\end{figure}

\subsection{Applications}
\paragraph{Versatile Editing}
Our model exhibits strong generalization ability, allowing seamless integration with outputs from advanced image generation or editing models.
As illustrated in Fig.~\ref{fig: edit}, it can transform images synthesized by text-to-image or image-editing models into fully animatable 3D avatars, enabling flexible downstream editing and control.

\paragraph{Multiple Inputs}
Recent works~\cite{pflhm} attempt to handle multi-view human images using computationally alternative attention~\cite{vggt}.
In contrast, our method can naturally handle multi-view inputs by simply linearly blending the UV features from each image.
As illustrated in Fig.~\ref{fig: multi}, this simple strategy is sufficient to produce coherent reconstructions even in challenging multi-view scenarios.

\subsection{Ablation Study}
\paragraph{Model Design}
We ablate the decoder architecture, as summarized in Tab.~\ref{tab: ablation} (a) and (d).
Increasing the number of decoder layers leads to a consistent performance gain.
In addition, introducing the shell token provides a clear boost, indicating that the extra token effectively enriches local surface modeling and, in turn, improves reconstruction quality.

\paragraph{Dataset Scalability}
We also study the impact of training data type and scale.
As shown in Tab.~\ref{tab: ablation} (b) and (c), model performance improves steadily as the dataset grows, highlighting the benefit of larger and more diverse supervision.

When trained only on synthetic data \textit{syn}, the model shows limited generalization.
Using only generation data \textit{gen} alleviates this issue, while $\textit{gen}^{*}$, which contains only video-generated data without augmented images, performs worse.
The best results are achieved when combining all data sources, suggesting that accurate 3D simulations complemented by diverse 2D views are crucial for high-quality reconstruction.

\begin{table}[t]
\centering
\caption{
Ablation study on model design and dataset scalability, evaluated on upper-body synthetic clips.
\colorbox{blue!50}{Blue} indicates the best results.
}

\label{tab: ablation}

\vspace{-4mm}

\makebox[\linewidth]{%
\subcaptionbox{%
Decoder depth}{
    \scalebox{0.7}{
    \begin{tabular}{cccc}
    \toprule
    blocks & PSNR↑ & SSIM↑ & LPIPS↓ \\ \midrule
    2 & 20.31 & 0.7846 & 0.1812 \\
    4 & 20.36 & 0.7862 & 0.1789 \\
    8 & \cb{20.38} & \cb{0.7867} & \cb{0.1635} \\
    \bottomrule
    \end{tabular}
    }
}

\subcaptionbox{%
Dataset scale}{
\scalebox{0.7}{
    \begin{tabular}{cccc}
    \toprule
    ratio       & PSNR↑ & SSIM↑ & LPIPS↓ \\ \midrule
    $95\%$     & \cb{20.38} & \cb{0.7867} & \cb{0.1635} \\
    $33\%$      & 20.33 & 0.7855 & 0.1801 \\
    $3\%$       & 20.25 & 0.7827 & 0.1888 \\
    \bottomrule
    \end{tabular}
    }
}

}

\makebox[\linewidth]{%
\subcaptionbox{%
Dataset type}{
    \scalebox{0.7}{
    \begin{tabular}{cccc}
    \toprule
    type       & PSNR↑ & SSIM↑ & LPIPS↓ \\ \midrule
    \textit{full}       & \cb{20.38} & \cb{0.7867} & \cb{0.1635} \\
    \textit{syn}        & 19.57 & 0.7579 & 0.2418 \\
    \textit{gen}        & \cb{20.38} & 0.7861 & 0.1732 \\
    \textit{gen$^*$}    & 20.34 & 0.7851 & 0.1794 \\
    \bottomrule
    \end{tabular}
    }
}
\hspace{0.15cm}

\subcaptionbox{%
Shell token}{
    \scalebox{0.7}{
    \begin{tabular}{cccc}
    \toprule
               & PSNR↑ & SSIM↑ & LPIPS↓ \\ \midrule
    \makecell[c]{\footnotesize \textit{w/} \\ \footnotesize \textit{shell token}}       & \cb{20.38} & \cb{0.7867} & \cb{0.1635} \\
    \makecell[c]{\footnotesize \textit{w/o} \\ \footnotesize \textit{shell token}} & 20.35 & 0.7845 & 0.1801 \\
    \bottomrule
    \end{tabular}
    }
}
}

\vspace{-5mm}
\end{table}
\section{Conclusion}
\label{sec:con}

We present a unified framework for reconstructing animatable 3D human avatars, spanning head-only, half-body, and full-body inputs within a single model. We train our model entirely on synthetic data, yet it generalizes well to in-the-wild portraits. Extensive experiments across head, upper-body, and full-body benchmarks demonstrate state-of-the-art or competitive performance, along with strong novel view synthesis and versatile applications such as reenactment, appearance editing, and multi-view fusion.

Despite these advances, our framework still has limitations. It relies on a proxy mesh that may fail in extreme or highly articulated poses, leading to noticeable reconstruction errors. Although our synthetic data manifold provides diverse identities and appearances, the views are still sparse, with limited side or rear perspectives, resulting in incomplete viewpoint coverage during training.

Future work will explore weakly pose-dependent representations to mitigate these issues and further enhance robustness and generalization. We also believe that improving viewpoint coverage and reducing the dependence on proxy geometry could further strengthen performance in more unconstrained real-world scenarios.

\section*{Acknowledgements}
The authors would like to express their sincere gratitude to Paul McIlroy, Tadas Baltrusaitis, and Charlie Hewitt for generously providing the synthetic human rendering pipeline, which played a crucial role in this project. Additionally, we are deeply thankful to Ross Cutler’s team for their continuous support and insightful discussions, which greatly enhanced the quality of our work.

{
    \small
    \bibliographystyle{ieeenat_fullname}
    \bibliography{main}
}

\clearpage
\setcounter{page}{1}

\newcommand{\maketitlesupplementarywoteaser}{%
  \newpage
  \twocolumn[{%
    \centering
    \Large
    \textbf{\thetitle}\\
    \vspace{0.5em}Supplementary Material \\
    \vspace{1.0em}

    \includegraphics[width=\textwidth]{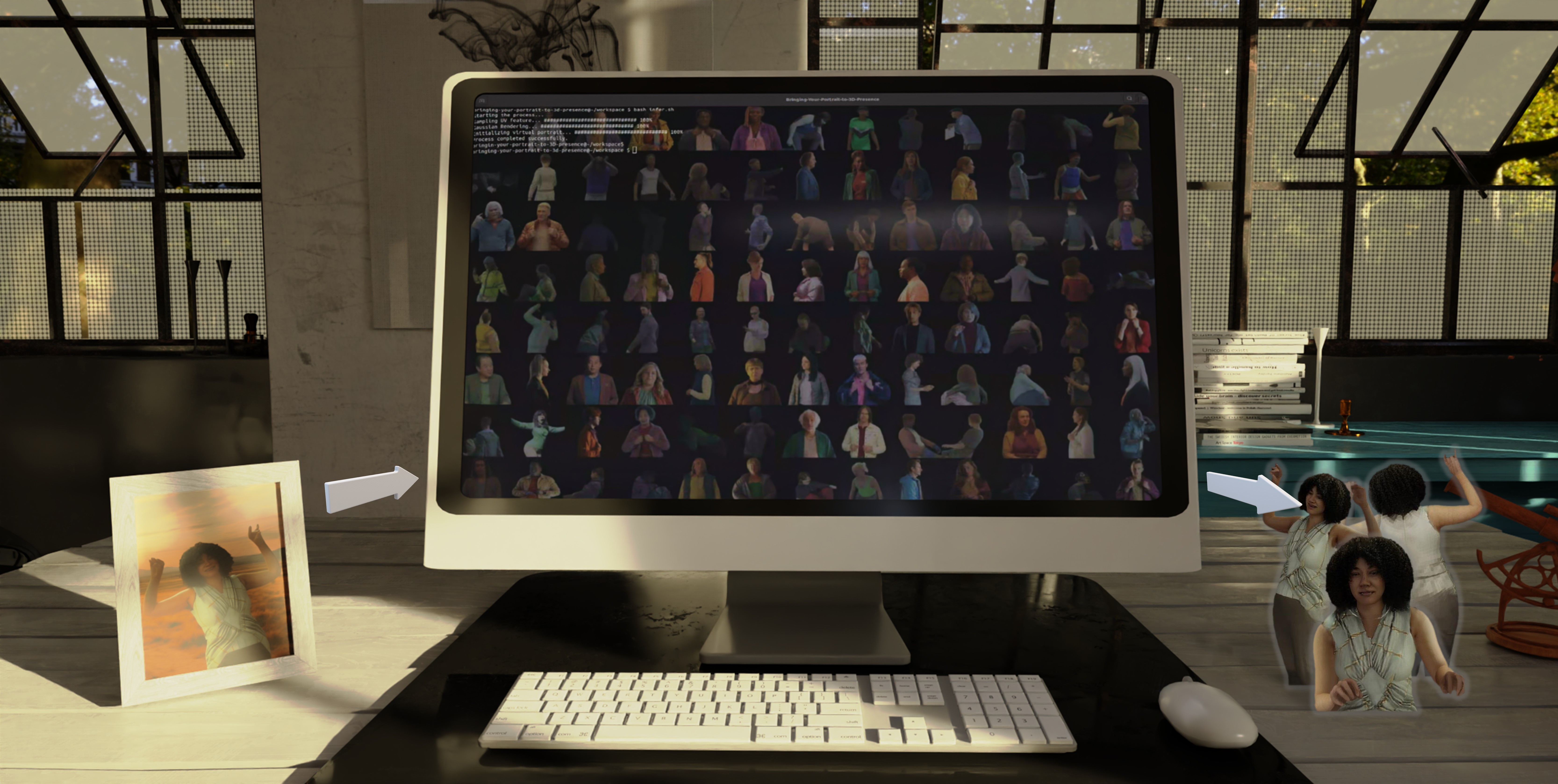}
    \captionof{figure}{A conceptual illustration of \textit{Bringing Your Portrait to 3D Presence}. Our pipeline transforms everyday portrait images into fully controllable 3D avatars that can be animated via a tracked proxy mesh. The model is trained entirely on a hybrid synthetic corpus combining rendered and generative sources. Thanks to our dual-UV representation, the system robustly handles inputs of varying completeness—ranging from head-only to half-body or full-body portraits—within a single unified framework.}
    \label{fig:supp_teaser}

    \vspace{1.0em}
  }]%
}

\maketitlesupplementarywoteaser
\appendix

We provide the model design and training procedure in Sec.~\ref{sec: supp_model}, the details of our dataset curation in Sec.~\ref{sec: supp_dataset},  the proxy-mesh estimation pipeline in Sec.~\ref{sec: supp_tracker}, and more experiments in Sec.~\ref{sec: supp_exp}. Additional qualitative results are shown in the supplementary videos.

\section{Model Details}
\label{sec: supp_model}

\subsection{Model Design}
\begin{figure}[]
    \centering
    \includegraphics[width=0.5\textwidth]{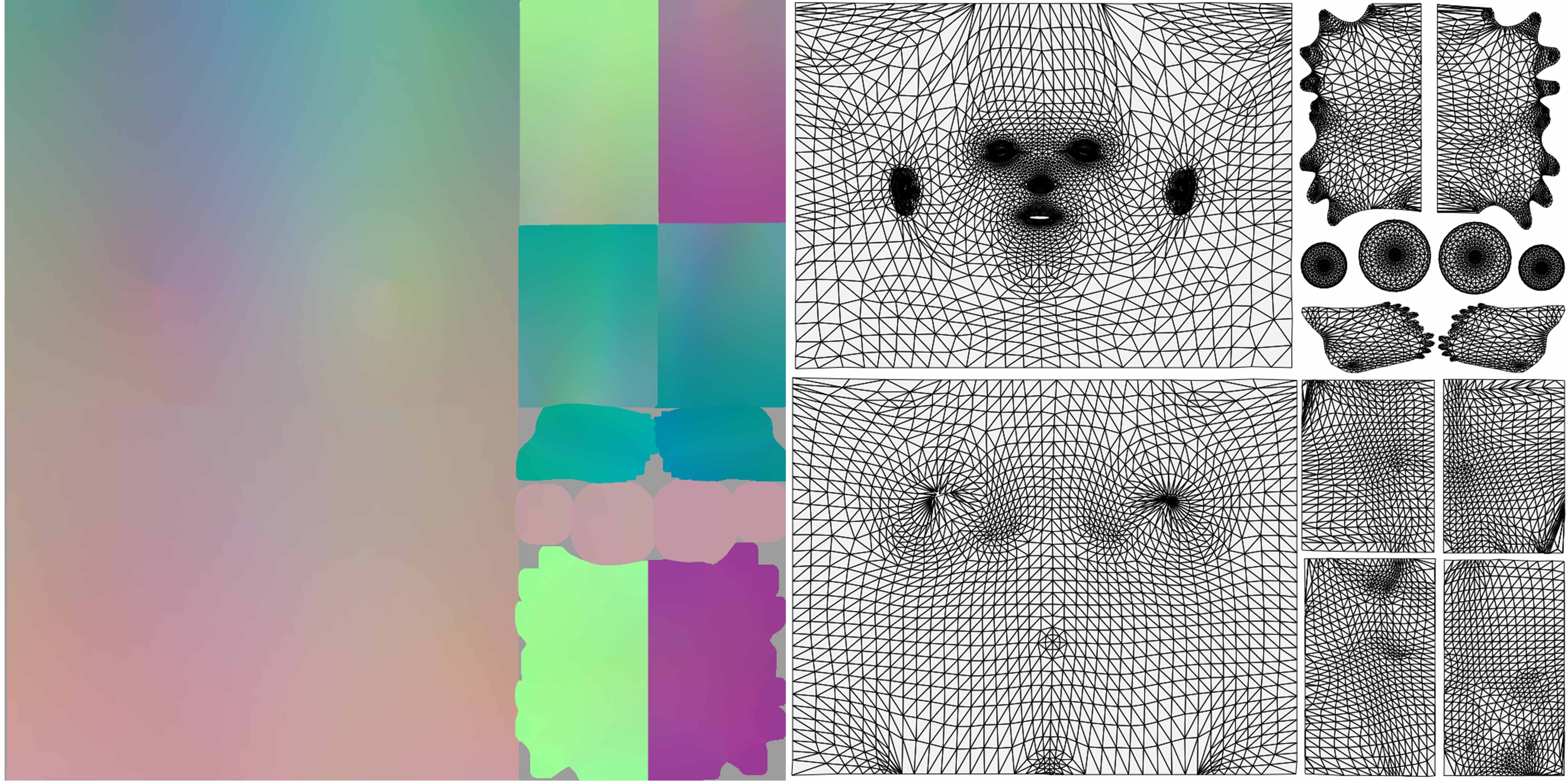}
    \vspace{-1em}
    \caption{\textbf{UV Topology Visualization and Position Map}.
    We visualize the modified UV topology and the corresponding position map used for sinusoidal encoding.
    }
    \label{fig: supp_uv}
    \vspace{-8px}
\end{figure}

After scattering UV features, we add separate learnable positional embeddings for different UV branches.
For the \emph{core-UV} branch, we initialize the positional embedding in UV coordinates.
Concretely, we rasterize the vertices of the canonical-space template mesh into the UV plane and obtain a position map shown in Fig.~\ref{fig: supp_uv}, with the same spatial resolution as the UV feature map.
It is worth noting that, although GSM~\cite{gsm} also adopts a shell-based design, our Shell-UV is different in design. While GSM adds extra 3D Gaussian layers for geometric expressiveness, our Shell-UV adds no Gaussians and instead uses canonical UV projection to reduce pose- and framing-induced misalignment under partial visibility.
We then apply an $L$-frequency sinusoidal encoding to each UV coordinate, with $L = 8$, and pass the encoded features through a linear layer to project them to the Sapiens feature dimension. For the \emph{shell-UV} branch, we initialize the learnable tokens with Gaussian noise.

The UV features with positional embeddings are linearly projected to 1024 and processed by $8$ self-attention blocks with $16$ attention heads each, to model interactions among tokens.
We then perform unpatchify with patch size $8$ to form Gaussian attribute maps.
For different Gaussian attributes, we use separate decoder heads, each implemented as a two-layer MLP with 256 hidden dimension and SiLU activation.
We uniformly sample from the decoded Gaussian attribute maps and slightly retopologize the UV layout so that the utilization of tokens is as high as possible.

During training, we randomly mask from 0$\%$ to 50$\%$ of the scattered UV features to improve robustness. At test time, we use the input image mask to filter out points that fall outside the masked region due to mesh misalignment. This simple trick effectively prevents artifacts caused by proxy-mesh misalignment.

\subsection{Loss Function}
Given a reference–target pair, we predict canonical Gaussian attributes from the reference, rig them to the target mesh, and render both views. 
The total objective combines reconstruction and regularization terms:
\begin{equation}
\mathcal{L}=\mathcal{L}_{\mathrm{rec}}+\mathcal{L}_{\mathrm{regu}}.
\end{equation}

\noindent\textbf{Reconstruction loss.}
For each view $v\!\in\!\{\mathrm{ref},\mathrm{tgt}\}$, we supervise image fidelity using pixel and perceptual losses:
\begin{equation}
\mathcal{L}_{\mathrm{rec}}^{(v)} 
=\lambda_{\mathrm{L}1}\!\left\|\hat{I}^{(v)}-I^{(v)}\right\|_1
+\lambda_{\mathrm{LPIPS}}\,
\mathcal{L}_{\mathrm{LPIPS}}\!\left(\hat{I}^{(v)},I^{(v)}\right),
\end{equation}
with $\lambda_{\mathrm{L}1}=1.0$ and $\lambda_{\mathrm{LPIPS}}=0.5$.  
The reconstruction loss sums over both views:
\begin{equation}
\mathcal{L}_{\mathrm{rec}}
=\mathcal{L}_{\mathrm{rec}}^{\mathrm{ref}}+\mathcal{L}_{\mathrm{rec}}^{\mathrm{tgt}}.
\end{equation}

\noindent\textbf{Geometry regularization.}
We constrain Gaussian geometry parameters to ensure stability.  
Offsets are penalized to prevent drift:
\begin{equation}
\mathcal{L}_{o}=\|\boldsymbol{d}\|_2,
\end{equation}
and scales $\boldsymbol{s}=[s_x,s_y,s_z]$ are regularized for compactness and isotropy:
\begin{align}
\mathcal{L}_{s}&=\sum_{i\in\{x,y,z\}} s_i,\\
\mathcal{L}_{r}&=\max\!\left(\frac{\max(\boldsymbol{s})}{\min(\boldsymbol{s})}-r,\,0\right),
\end{align}
where $r\!=\!9$.  
The geometry regularization term is
\begin{equation}
\mathcal{L}_{\mathrm{geo}}
=\lambda_o\mathcal{L}_o+\lambda_s\mathcal{L}_s+\lambda_r\mathcal{L}_r.
\end{equation}

\noindent\textbf{Texture regularization.}
We further regularize Gaussian appearance to avoid implausible color and opacity distributions. 
A hand-consistency loss aligns hand and facial color statistics by matching each hand patch $\mathbf{P}_i^h$ to its nearest facial patch $\mathbf{P}_j^f$ (gradient detached):
\begin{equation}
\mathcal{L}_{h}=\sum_i \min_j \|\mathbf{P}_i^h-\mathbf{P}_j^f\|_2.
\end{equation}

In addition, we impose a patch-level regularization on the opacity. For each Gaussian opacity map, we divide it into patches and enforce that the average opacity in each patch is close to a target value $\alpha_{\mathrm{ref}}$, which we set as 0.8. The opacity loss for each patch is defined as:
\begin{equation}
\mathcal{L} _{\alpha}=-\frac{1}{N}\sum_{k=1}^N{\left( \alpha _{\mathrm{ref}}\cdot \log \left( \mu _k \right) +(1-\alpha _{\mathrm{ref}})\cdot \log \left( 1-\mu _k \right) \right)},
\end{equation}
where $\mu_k$ is the mean opacity of the $k$-th patch, and $N$ is the number of patches. 

The texture regularization term is
\begin{equation}
\mathcal{L}_{\mathrm{tex}}
=\lambda_h\mathcal{L}_h + \lambda_{\alpha}\mathcal{L}_{\alpha}.
\end{equation}

\noindent Finally, the total regularization is
\begin{equation}
\mathcal{L}_{\mathrm{regu}}
=\mathcal{L}_{\mathrm{geo}} + \mathcal{L}_{\mathrm{tex}},
\end{equation}
with $(\lambda_o,\lambda_s,\lambda_r,\lambda_h,\lambda_{\alpha}) = (1.0, 0.1, 1.0, 0.1, 0.1)$ in all experiments.
\section{Dataset Details}
\label{sec: supp_dataset}

\subsection{Synthetic Rendering Branch}
Our synthetic rendering pipeline begins by sampling shape and pose parameters from the SOMA parametric human model~\cite{lookma}.
Given the sampled body and pose coefficients, we apply texture, hair and clothing assets upon posed mesh.
And further sample a HDR image to set up environment lightning.
The full scene is rendered using the Cycles rendering engine.
For camera configuration, we adaptively determine the look-at point based on the posed subject rather than using a fixed center.
Let $v_{\mathrm{min}}$ and $v_{\mathrm{max}}$ denote the minimum and maximum vertex positions in world space along the vertical axis, and let $\mathbf{p}_{\mathrm{pel}}$ and $\mathbf{p}_{\mathrm{head}}$ be the pelvis and head joint locations, respectively. We define the horizontal look-at target as
\begin{equation}
    \mathbf{t}_{xy}=\frac{\mathbf{p}_{\mathrm{pel}}+\mathbf{p}_{\mathrm{head}}}{2},
\end{equation}
and the vertical target coordinates as
\begin{equation}
    t_z=v_{\min}+\lambda \cdot \left( v_{\max}-v_{\min} \right) ,\quad \lambda =0.75.
\end{equation}
Thus, the final camera look-at position is $\mathbf{t}=\left( \mathbf{t}_{xy},t_z \right)$, which biases the view toward the upper-body and yields perceptually stable framing across diverse poses and body shapes.

\begin{figure*}[!htbp]
    \centering
    \includegraphics[width=0.9\textwidth]{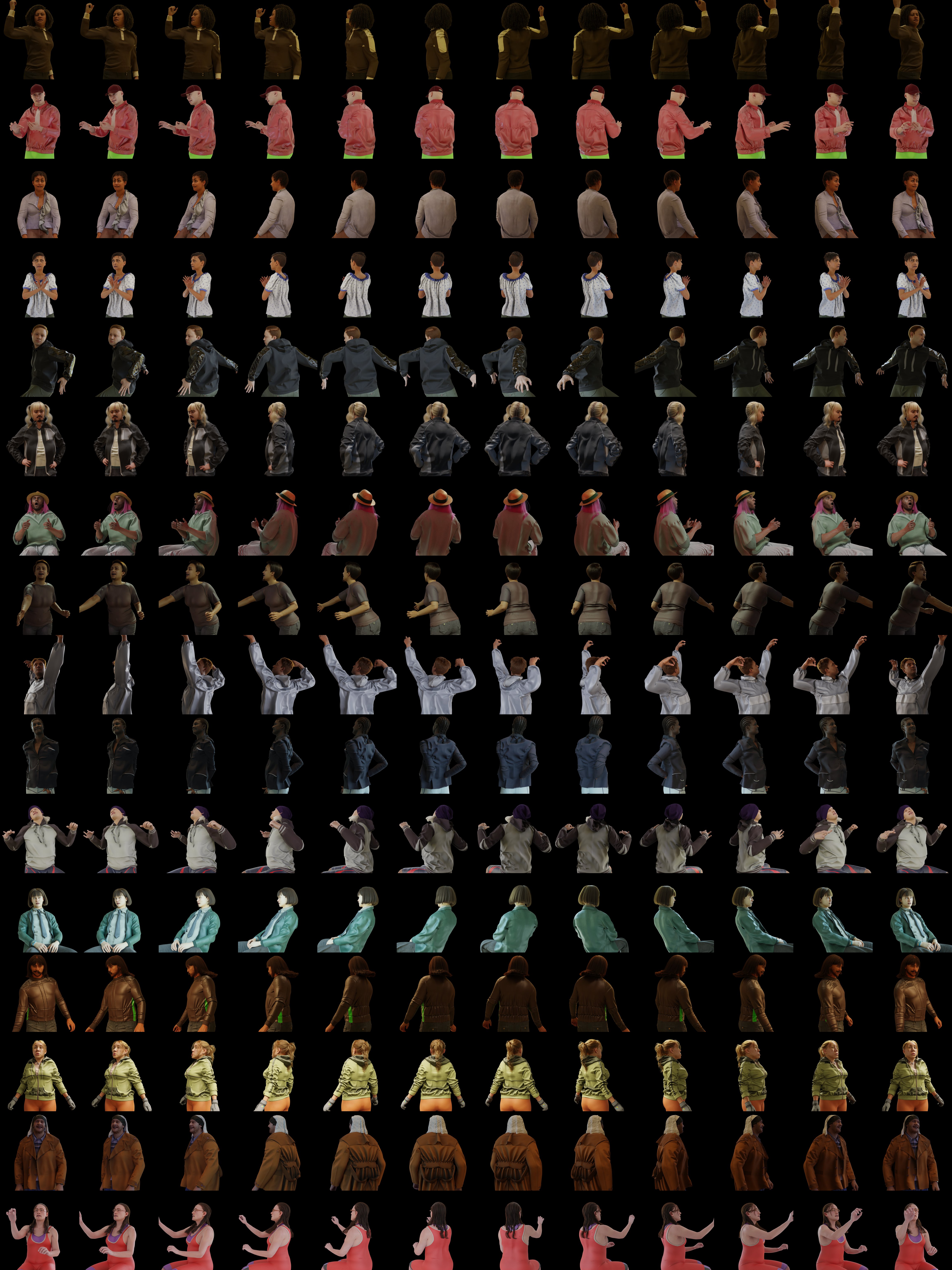}
    \caption{\textbf{Synthetic Rendering
    Dataset}.
    Our synthetic rendering dataset contains diverse body poses, rendered from multiple viewpoints with perfect mesh annotations, providing strong structural priors for model training.
    }
    \label{fig: supp_syn_data}
\end{figure*}

In addition to the rendered images, we export the corresponding SOMA mesh. We then convert it into an SMPL-X using a pre-computed regression matrix, followed by a parameter inversion step in which we optimize the SMPL-X parameters via trust-region Newton conjugate gradient method~\cite{NoceWrig06} to best match the converted mesh. We show samples from the rendered dataset in Fig.~\ref{fig: supp_syn_data}.

\subsection{Generative Branch}
\subsubsection{Semantic Factorization}

We construct a factorized latent space by decomposing the generative prior into semantically meaningful axes. For each factor, we curate a dedicated vocabulary that captures its underlying variation (e.g., appearance attributes, scene composition, or stylistic cues). These vocabularies collectively define controllable semantic directions along which data diversity can be systematically spanned. These vocabularies are automatically expanded using GPT-5 to ensure broad coverage while preserving the semantic purity of each factor.

\begin{itemize}[leftmargin=*]
    \item \textit{Actions} have 131 carefully constrained micro‑gestures grouped implicitly (speaking, explanatory, facial, grooming, fit‑check) and phrased as short dynamic–return patterns to stabilize temporal synthesis while avoiding occlusion or exaggerated motion.

    \item \textit{Hair} splits into 26 natural color variants and 54 face‑visible styles spanning length, texture, braids, locs, curls, and updos.

    \item \textit{Lighting control} uses 27 daylight and 25 night or practical scenario words plus 22 declarative reinforcement lines to couple physically plausible cinematography (direction, diffusion, key/fill balance, rim separation) with exposure stability (“Exposure anchors on the face”, “Shadow detail is preserved”).

    \item \textit{Outfit} is factored into orthogonal granularities: 64 balanced color tones (8 groups $\times$ 8 hues), 45 fabric/material descriptors, a large library of tops ($>$400) and outerwear ($>$1000) with an optional `None` sentinel for absence, plus 45 accessories likewise optionally omitted. Surface semantics are isolated into micro pattern sets (woven, knit, formal micro-structures), small neutral wordmarks, a large bank of mid‑scale front graphics ($>$300 non‑branded, stylized motifs), all‑over prints, color‑blocking schemes, embroidery/appliqué types, and extensive outerwear construction/detail modifiers ($>$200).

    \item \textit{Role} and attire archetypes form the largest axis ($>$1000 unique descriptors) spanning contemporary professions, protective gear, historical armors, global traditional garments, performing arts, subcultures, sports kits, speculative sci‑fi, fantasy and genre motifs, craft and maritime occupations, emergency and technical variants—each phrase bundling a silhouette anchor plus distinctive accessories for high visual discriminability while remaining culturally neutral.

    \item  \textit{Subjects} provide 24 age granularity words (coarse bands plus early/mid/late decades), 52 region‑level origin abstractions (continental and sub‑regional without nationality specificity), and inclusive gender nouns (“man”, “woman”, “person”, “non-binary person”), intentionally broad to mitigate bias.

    \item \textit{Time} contributes 26 day/night or twilight states (“golden hour”, “civil twilight after sunset”, “dawn blue hour”) paired with a coarse day/night flag, directly complementing lighting vocabulary to steer chromatic and contrast regimes.
    
\end{itemize}

\subsubsection{Scene Composition}

Given the factorized vocabularies defined above, we next compose them into complete scene descriptions that serve as conditioning signals for the upper-body video generator. Rather than relying on free-form textual prompts, we programmatically assemble each scene by sampling a small set of descriptors along the active semantic axes—such as subject identity cues, actions, hairstyle, lighting setup, outfit or role, and time-of-day—and inserting them into a structured scene template. This approach ensures that every generated instance is grounded in the same underlying factor space while still exhibiting rich and controlled visual diversity.

To further broaden the appearance distribution, we introduce two complementary scene-composition regimes that differ in how clothing-related factors are instantiated. These regimes share the same semantic axes but emphasize distinct clothing conventions, allowing us to explore a wider range of apparel variability without altering the core factorization.

\begin{itemize}[leftmargin=*]
    \item \textbf{Outfit-centric composition.} 
    In this regime, the scene description is constructed by foregrounding the explicit outfit-related vocabulary—including color palettes, fabric types, garment categories (e.g., tops, outerwear), accessories, and surface patterns—while deliberately leaving the ``role’’ attribute unspecified. This strategy encourages the generator to synthesize visually clean and relatively simple garments that are easier to segment, normalize, and analyze. It also provides more disentangled control over low-level appearance attributes by isolating clothing factors from higher-level semantic cues. The corresponding outfit-centric template is shown below:
    
    \begin{quote}\small
    \emph{\{time\_of\_day\}, \{lighting\}, \{shot\_size\}, center composition. 
    A/An \{age\} \{gender\_noun\} (from \{region\}) wearing a \{top\_color\} \{top\_fabric\} \{top\} 
    with/featuring \{top\_decoration\} paired with a \{outer\_color\} \{outerwear\} \{outerwear\_detail\}
    and \{accessory\}, with \{hair\_color\} \{hairstyle\}. 
    The person \{action\}, in a waist-up, standing, fixed-camera shot with arms and hands visible; lighting remains stable and physically plausible.}
    \end{quote}
        
    \item \textbf{Role-centric composition.} 
    In this regime, the scene description is anchored on the rich role–attire archetype axis, which provides high-level cues about profession, social persona, cultural style, or situational context. Once a role is selected, the remaining factors—such as subject attributes, action, hairstyle, lighting, and time-of-day—are sampled to remain semantically compatible with the chosen archetype. Because each role phrase implicitly encodes a characteristic silhouette, associated accessories, and distinctive detailing, this strategy naturally produces outfits with more elaborate structure and heightened stylistic diversity compared with the outfit-centric scheme.

    \begin{quote}\small
    \emph{\{time\_of\_day\}, \{lighting\}, \{shot\_size\}, center composition.
    A/An \{age\} \{gender\} \{role\} from \{region\}, wearing characteristic \{role\}-specific clothing and accessories, with \{hair\_color\} \{hairstyle\}. 
    The person \{action\}, in a waist-up, standing, fixed-camera shot with arms and hands visible; lighting remains stable and physically plausible.}
    \end{quote}
    
\end{itemize}

A corresponding negative description is constructed in the same manner by sampling several terms from a curated list of undesired artifacts (e.g., low resolution, motion jitter, flicker, over- or under-exposure, extreme torso crops, seated or occluded poses). Aside from this auxiliary negative specification, the two scene-composition strategies share an identical sampling pipeline; they differ solely in how clothing-related information is selected, emphasized, and integrated into the final scene description.

\subsection{Filmic Realism Regularization}

The factorized samplers introduced above produce scene descriptions that are structurally clean and fully disentangled across semantic axes, but the resulting text is intentionally minimal. In the outfit-centric regime, such terseness is acceptable: the specification primarily consists of low-level, compositional attributes—colors, fabrics, garments, simple actions—and can already drive the generator to produce plausible videos. In contrast, the role-centric regime operates at a much higher semantic level. A single role indicator implicitly encodes equipment, safety constraints, cultural context, and a characteristic visual grammar. Directly combining these role cues with independently sampled attributes often results in descriptions that are grammatically valid yet visually implausible or internally inconsistent (e.g., ``a firefighter in full bunker gear with long, loose hair flowing over the shoulders'').

\begin{figure*}[!htbp]
    \centering
    \includegraphics[width=1.0\textwidth]{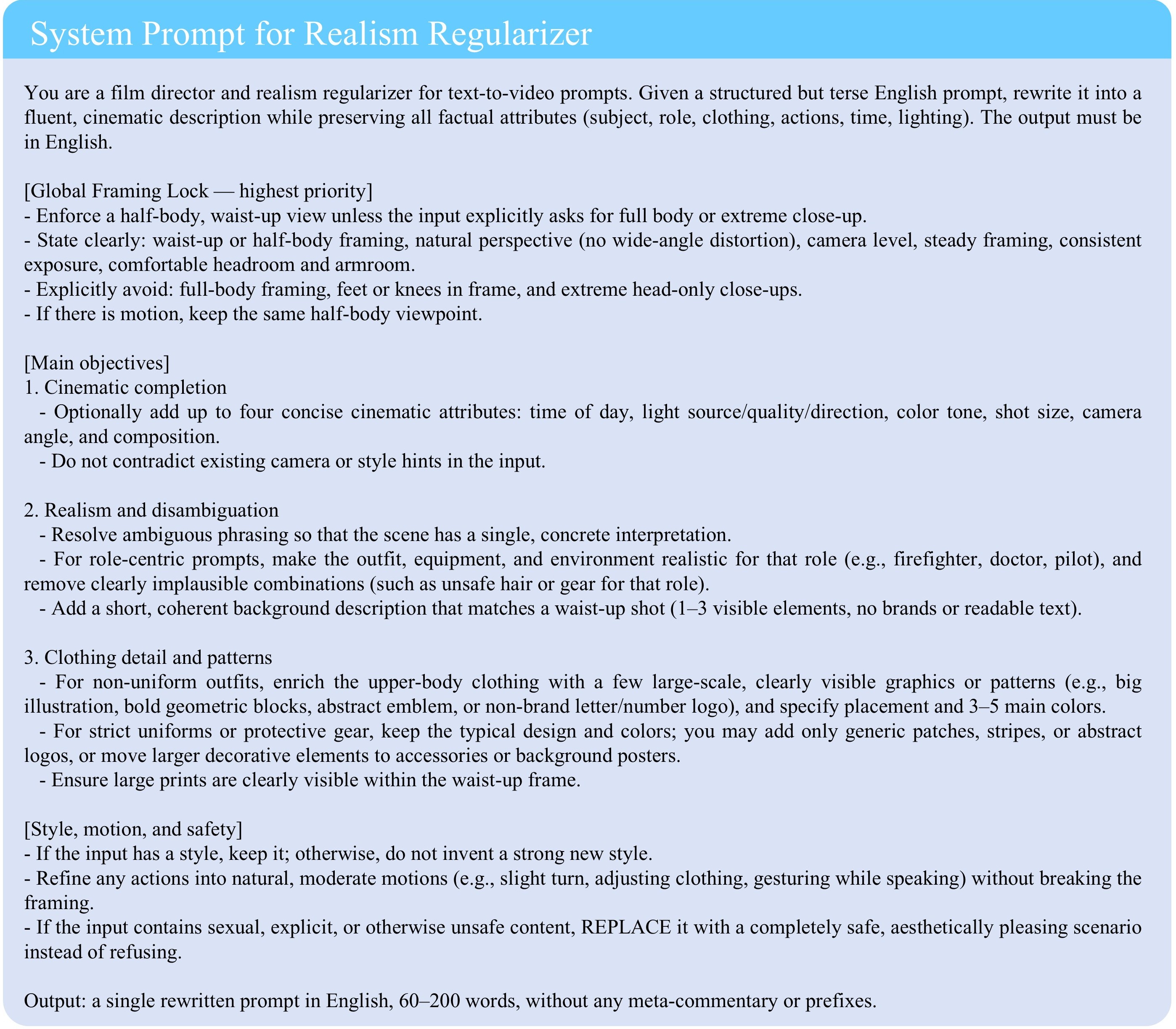}
    \vspace{-1em}
    \caption{\textbf{Filmic Realism Regularization}.
   The structured templates are processed by a lightweight LLM that improves linguistic fluency and resolves inconsistencies, yielding scene descriptions with enhanced realism and contextual coherence.
    }
    \label{fig: prompt_extension}
    \vspace{-8px}
\end{figure*}

This mismatch highlights a key insight in our data design: \emph{semantic factorization alone does not guarantee filmic coherence}. High-level roles impose structured dependencies among appearance, action, accessories, and physical context—dependencies that must be restored for the compositions to resemble real-world footage. To address this, we leverage a lightweight language-model-based postprocessor that acts as a \textit{filmic realism regularizer}. Given a structured template as input, the instruction-tuned Qwen2.5-14B-Instruct~\cite{qwen25} rewrites the description into a fluent, naturalistic scene while preserving all controllable factors introduced by the sampler. As illustrated in Fig.~\ref{fig: prompt_extension}, the model performs three key types of corrections:

\begin{itemize}[leftmargin=*]
    \item \textbf{Disambiguation.} Remove or resolve ambiguous phrasing in the compositional template (e.g., clarifying vague actions or lighting descriptions) so that a single, concrete visual interpretation is implied.
    \item \textbf{Role-aware scene completion.} For role-centric compositions, insert an appropriate surrounding scene or objects that are compatible with the specified role (e.g., adding a station, workplace, or tools) and remove attribute combinations that contradict typical equipment or safety requirements.
    \item \textbf{Richer clothing detail.} Elaborate the clothing description with additional but compatible details and surface patterns (e.g., stitching, pockets, insignia, emblems), increasing visual complexity without changing the underlying factors selected by the sampler.
\end{itemize}

\subsection{Video Generation}
\begin{figure*}[]
    \centering
    \includegraphics[width=0.9\textwidth]{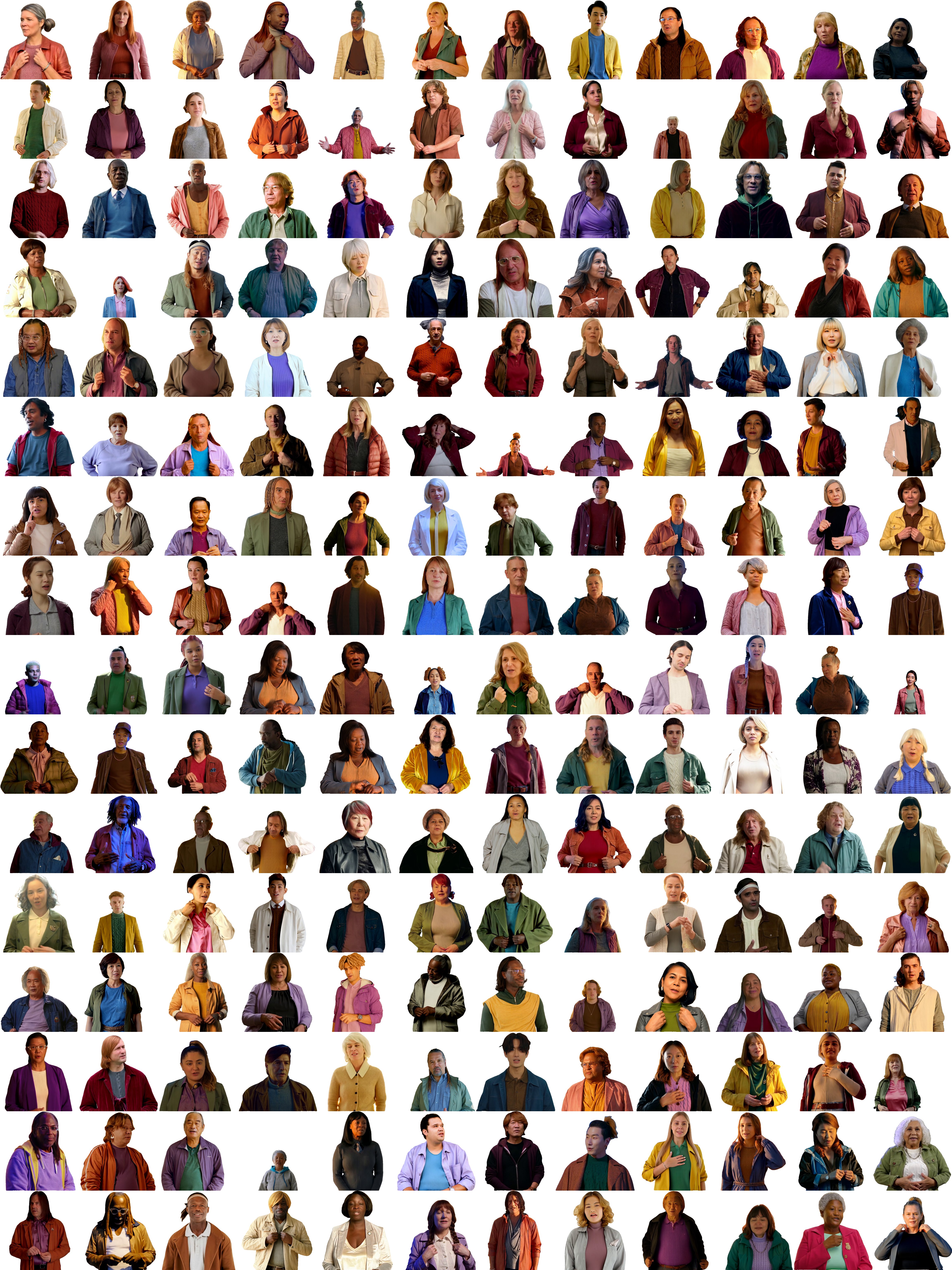}
    \caption{\textbf{Outfit-centric Generation}.
    Generation guided by outfit produces visually coherent and structurally consistent human images.
    }
    \label{fig: supp_outfit_centric}
\end{figure*}

\begin{figure*}[]
    \centering
    \includegraphics[width=0.9\textwidth]{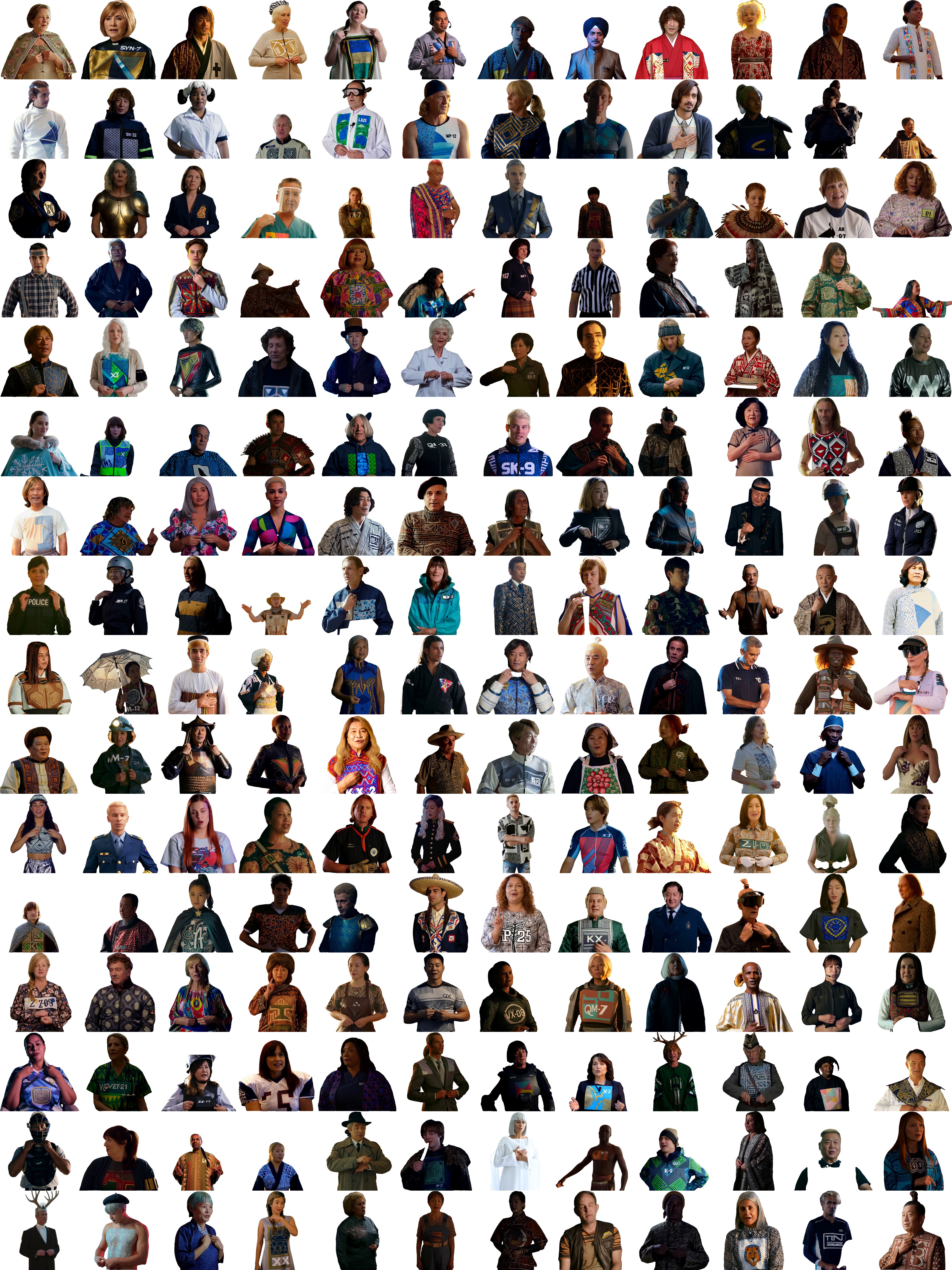}
    \caption{\textbf{Role-centric Generation}.
    Role-guided composition produces human images with noticeably more complex textures and styles.
    }
    \label{fig: supp_role_centric}
\end{figure*}

The refined scene descriptions are then fed into the Wan2.2-TI2V-5B~\cite{wan22} model, using its default inference configuration to synthesize upper-body video clips. The generator directly produces short sequences that inherit both the structured control from our factorized sampler and the filmic coherence enforced by the realism regularizer. Representative results for the two composition regimes are shown in Fig.~\ref{fig: supp_outfit_centric} and Fig.~\ref{fig: supp_role_centric}, which display the masked and cropped outputs for the outfit-centric and role-centric settings, respectively.

\subsection{Side/Back View Completion}
In practice, Wan2.2-TI2V-5B does not always produce stable, identity-consistent samples when a single person turns in place or rotates themselves. To supplement the multi-view supervision, we therefore perform simple side/back-view synthesis with Qwen-Image-Edit~\cite{qwenimage}: for each generated clip, we randomly sample one frame as the reference image and ask the model to generate left, right, and back views of the same subject. We use the following prompts for side and back views:

\begin{quote}\small
\textbf{Side view:}
\emph{``Change the subject to a left $90^\circ$ pure side profile (camera-left, yaw$\approx+90^\circ$).
Preserve identity (facial proportions/shape), hair length \& color, clothing color \& material, body height \& build;
keep lighting direction/intensity consistent; do not change the background or composition.
Photo-realistic, high resolution.''}

\vspace{0.4em}
\textbf{Back view:}
\emph{``Without changing the person's identity or composition, rotate the subject to a back-facing viewpoint ($\approx180^\circ$).
Preserve height, body shape, hair length \& color, clothing color \& material, and accessory positions;
keep lighting direction/intensity consistent; maintain the background's texture and perspective as much as possible.
The back view should be physically consistent with the front (collar shape, fabric folds, hair volume, shoulder line).
Photo-realistic, high resolution.''}
\end{quote}

\begin{figure*}[!htbp]
    \centering
    \includegraphics[width=0.9\textwidth]{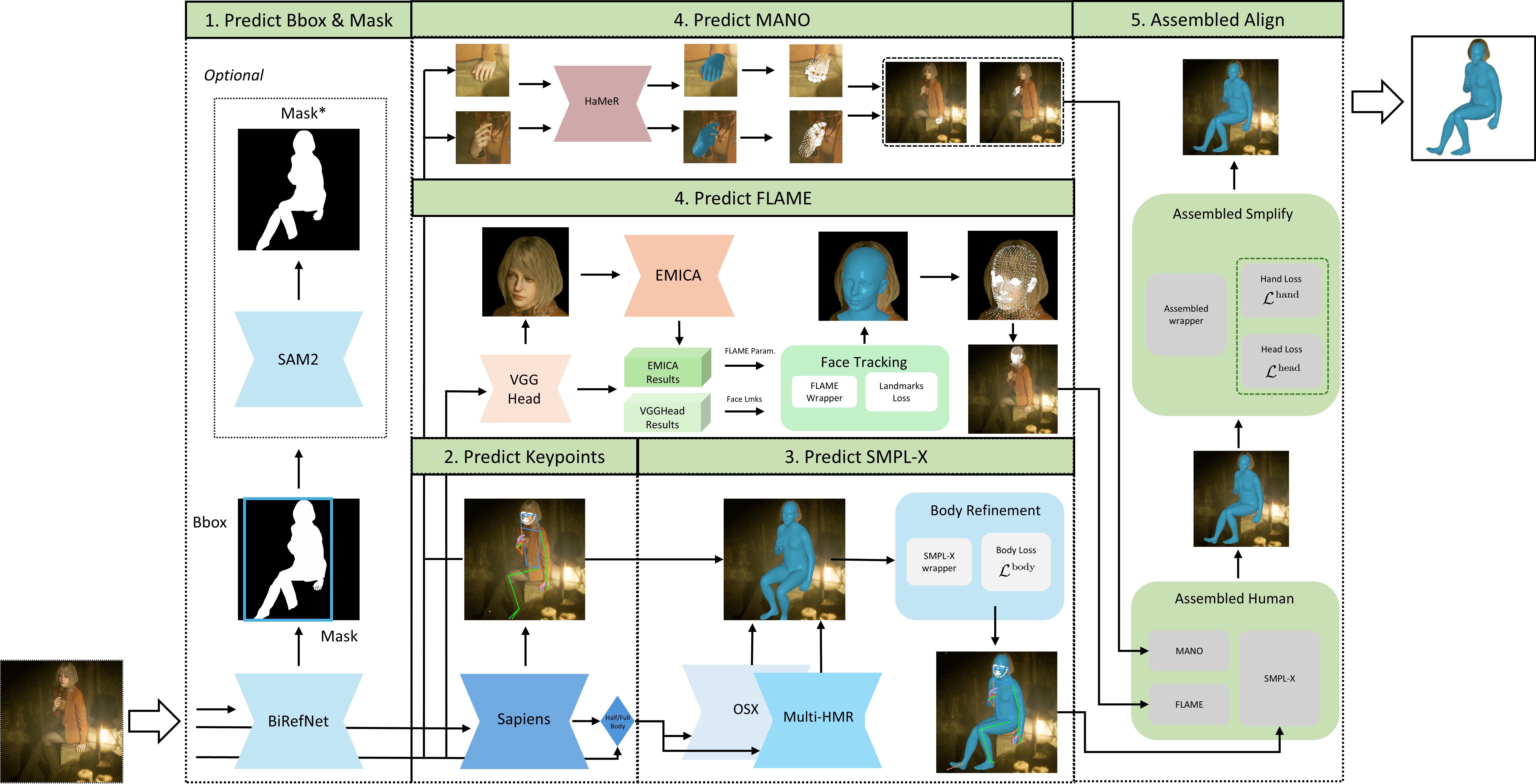}
    \caption{\textbf{Estimation Pipeline Diagram}.
    We illustrate our proxy-mesh estimation pipeline using a single image for clarity, while noting that the pipeline naturally supports parallel processing for multi-frame inputs. Starting from an input image, we preprocess it to extract a foreground mask and apply a pretrained human mesh recovery model to obtain an initial mesh estimate. The initial estimate is subsequently refined through body, head, and hand refinement.
    }
    \label{fig: supp_tracker_diagram}
    \vspace{-8px}
\end{figure*}

\begin{figure*}[]
    \centering
    \includegraphics[width=0.9\textwidth]{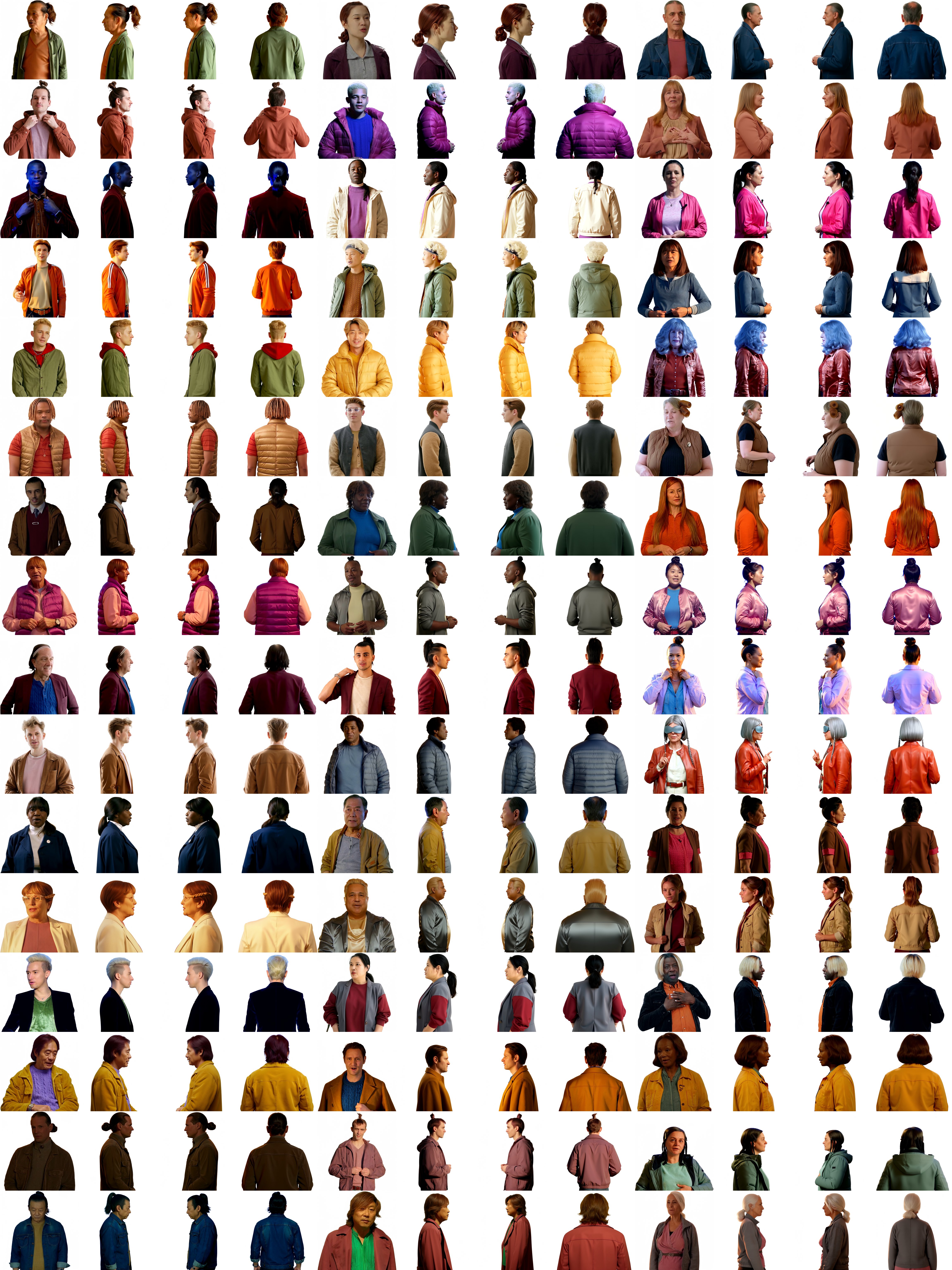}
    \caption{\textbf{Side/Back-view Augmentation}.
    We leverage advanced image-editing models to supplement abundant side- and rear-view information.
    }
    \label{fig: supp_qwen_edit}
\end{figure*}

Guiding an image-editing model to rotate a person to a side-view using textual prompts is not always reliable. We empirically find that the success rate can be significantly improved by adopting a simple strategy: we always use a single side-view prompt, and obtain right-view samples by horizontally flipping the reference image before editing and flipping the edited result back afterward. This allows both left and right side views to be generated using the same textual prompt. The augmented dataset is illustrated in Fig.~\ref{fig: supp_qwen_edit}. Although the resulting poses are not strictly consistent with those in the input images, this mismatch is acceptable because our training is formulated over reference–target pairs.

\subsection{Discussion}
Our data pipeline is intentionally simple, and the model design remains lightweight, leaving substantial room for future enhancement. Because the realistic-style supervision views in our synthetic corpus mainly cover side and back angles, side-view supervision may cause Gaussians on the supervised side to attenuate due to projections from the opposite side. The generative data may also produce imperfect hand geometry, occasionally resulting in floating points around the fingers. In addition, the projection-based scattering and decoder inpainting introduce a natural transition between observed and unobserved regions. These aspects point to several promising directions, such as enriching supervision viewpoints, improving hand priors, or incorporating GAN-based or DPT-style refinement for smoother cross-view consistency. Nevertheless, even with this minimalistic design, our synthetic data pipeline and dual-UV model deliver faithful reconstructions, robust generalization across head/half/full-body inputs, and plausible novel-view synthesis, achieving state-of-the-art or highly competitive performance.
\section{Details of Proxy Mesh Estimation}
\label{sec: supp_tracker}

In the following, we introduce our proxy mesh estimation pipeline, which we will refer to as \textit{the tracker}.
The overall workflow of our tracker is illustrated in Fig.~\ref{fig: supp_tracker_diagram}.
Our tracker is designed to produce stable estimates for less-restricted inputs.
We use SMPL-X to represent the full body, FLAME to represent the head, and MANO to parameterize hands.

For SMPL-X, we parameterize the body with
shape coefficients $\boldsymbol{\beta}^{\text{smplx}} \in \mathbb{R}^{N_\beta}$,
expression coefficients $\boldsymbol{\psi}^{\text{smplx}} \in \mathbb{R}^{N_\psi}$,
pose coefficients
\[
\boldsymbol{\theta} =
\left[
\boldsymbol{\theta}^{\text{glob}},
\boldsymbol{\theta}^{\text{body}},
\boldsymbol{\theta}^{\text{lhand}},
\boldsymbol{\theta}^{\text{rhand}},
\boldsymbol{\theta}^{\text{jaw}}
\right]
\in \mathbb{R}^{3K},
\]
and a global translation vector $\mathbf{t} \in \mathbb{R}^3$.
For FLAME, we use shape coefficients $\boldsymbol{\beta}^{\text{flame}} \in \mathbb{R}^{N_\beta}$, expression coefficients $\boldsymbol{\psi}^{\text{flame}} \in \mathbb{R}^{N_\psi}$, and head pose coefficients $\boldsymbol{\theta}^{\text{flame}} \in \mathbb{R}^{3K_{\text{flame}}}$.
For MANO, we denote the MANO shape coefficients by
$\boldsymbol{\beta}^{\text{mano}} \in \mathbb{R}^{N_\beta^{\text{hand}}}$,
the hand pose coefficients by
$\boldsymbol{\theta}^{\text{mano}} \in \mathbb{R}^{3K_{\text{hand}}}$,
and the hand translation by
$\mathbf{t}^{\text{mano}} \in \mathbb{R}^3$.
In practice, MANO is instantiated separately for the left and right hands
(with parameters
$\boldsymbol{\theta}^{\text{mano,l}}$,
$\boldsymbol{\theta}^{\text{mano,r}}$, etc.).
Given a subject, we share the shape across time and models, i.e., all frames of the same video share the same SMPL-X and FLAME shape coefficients.

Previous trackers exhibit limitations under our setting.
The LHM tracker cannot capture expressive facial motion and performs poorly for upper-body or shoulder-up crops.
The GUAVA tracker achieves accurate results under the strict assumption that both hands and face are visible.
However, when hands are occluded or truncated, GUAVA tracker often produces unstable and unpredictable estimates.

\begin{figure*}[!htbp]
    \centering
    \includegraphics[width=0.85\textwidth]{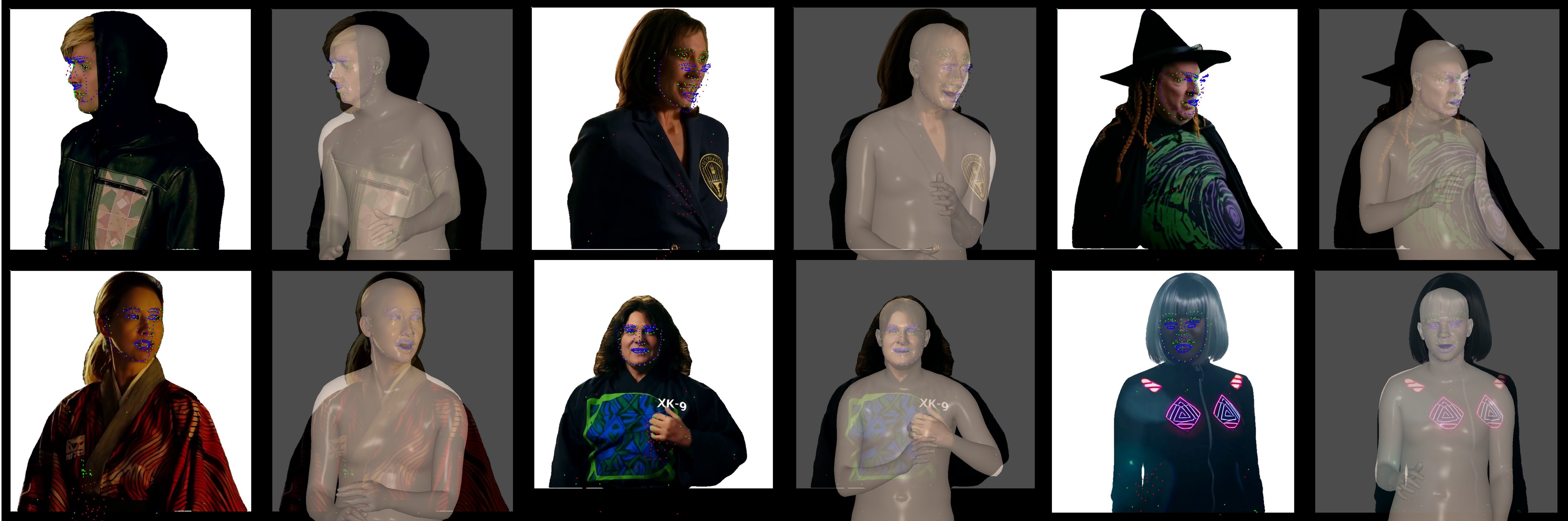}
    \caption{\textbf{Hands Missing Prediction}.
    Multi-stage methods, such as PIXIE, often produce unpredictable results when hand regions are missing.
    }
    \label{fig: supp_pixie_bad}
\end{figure*}

\begin{figure*}[!htbp]
    \centering
    \includegraphics[width=0.85\textwidth]{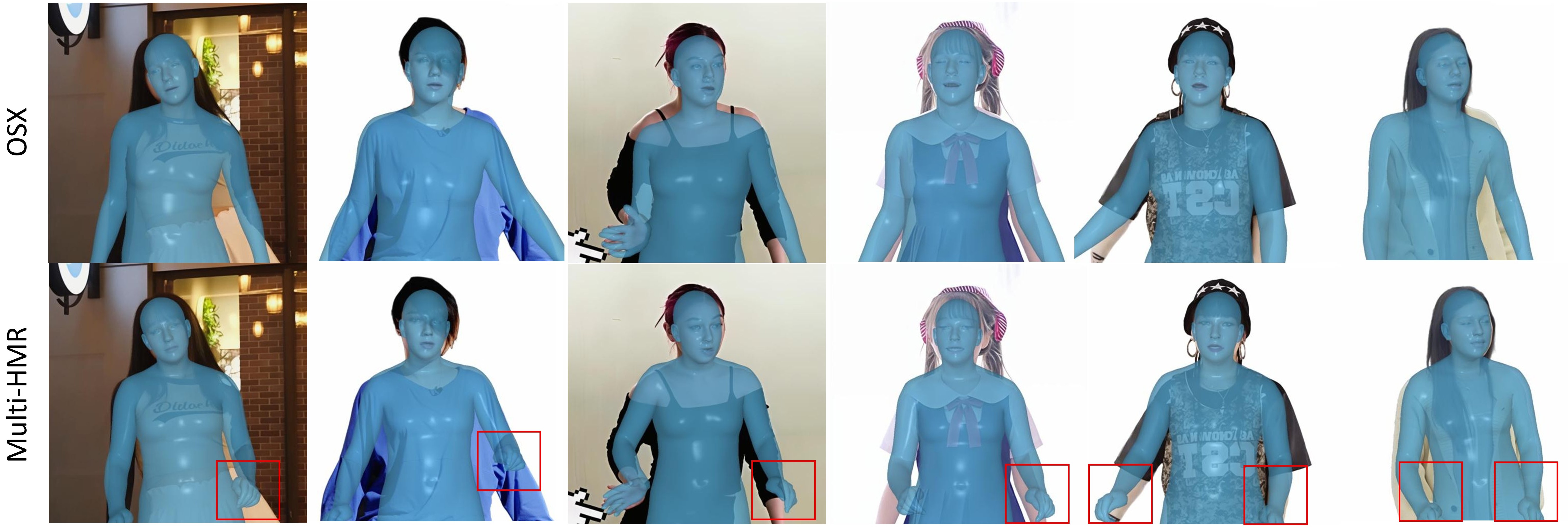}
    \caption{\textbf{Multi-HMR and OSX}.
    We find that OSX, trained primarily on upper-body data, produces reasonable results when hands are not visible, whereas MultiHMR often yields unsatisfactory predictions.
    }
    \label{fig: supp_multi_hmr_osx}
\end{figure*}

\subsection{Initial Estimates}
We first consider off-the-shelf human mesh recovery models for obtaining an initial solution from the input frame.
PIXIE~\cite{PIXIE} is a classic multi-stage model that crops out the face, hands, and body and processes them with dedicated encoders, whose features are then jointly fused.
However, when one or both hands are not visible, its hand estimates become highly unreliable as shown in Fig.~\ref{fig: supp_pixie_bad}, which also explains why GUAVA focuses on frames where both hands are visible.

In contrast, ViT-based methods such as Multi-HMR~\cite{multihmr} do not require explicit face or hand crops.
Nevertheless, in frames where the hands are not visible, they often default to placing the hands near the bottom of the image, presumably due to dataset bias.
OSX~\cite{osx}, a one-stage model trained on upper-body data, does not suffer from this issue.
This behavior is demonstrated in Fig.~\ref{fig: supp_multi_hmr_osx}.
Therefore, in our tracker we implement both Multi-HMR and OSX: Multi-HMR is used for full-body inputs, while OSX is used for upper-body inputs.
Since our curated dataset mostly contains upper-body views, we mainly use OSX in preprocessing.

\subsection{FoV Correction}
Methods such as OSX are trained under a very narrow FoV assumption ($\approx\!2^{\circ}$), which approximates the perspective projection with an orthographic one.
While this is helpful for recovering a stable 3D pose from a single image, it introduces large uncertainty when anchoring 3D Gaussians on the SMPL-X mesh.
Given an input frame, after obtaining an initial SMPL-X estimate from OSX, we first convert the camera intrinsics and the mesh translation along the $z$-axis to a canonical FoV of $30^\circ$.

Let $(f_x, f_y)$ denote the original focal lengths and $c_x$ the principal point in the horizontal direction.
We compute a new focal length $f_x'$ based on the desired FoV and define the scaling factor
\begin{equation}
s_x = \frac{f_x'}{f_x}.
\end{equation}
To preserve the aspect ratio, we set $f_y' = s_x f_y$.
We also update the depth translation using the same scale, i.e., $t_z' = s_x t_z$.

\subsection{Optimization}
\paragraph{Body Refinement.}
Using this canonicalized camera, we back-project Sapiens 2D keypoints to the 3D mesh and optimize the global orientation $\boldsymbol{\theta}^{\text{glob}}$, body pose $\boldsymbol{\theta}^{\text{body}}$, hand poses $\boldsymbol{\theta}^{\text{lhand}}, \boldsymbol{\theta}^{\text{rhand}}$, and translation $\mathbf{t}$.
This first stage aligns the SMPL-X mesh to the input frame with the following objective:
\begin{equation}
\begin{aligned}
\mathcal{L}^{\mathrm{body}} = {} &
\lambda_{\mathrm{reproj}} \, \mathcal{L}_{\mathrm{gmof}}\!\left(K, \hat{K}\right)
+ \lambda_{\mathrm{mask}} \, \mathcal{L}_{\mathrm{mask}}\!\left(M, \hat{M}\right) \\
& + \lambda_{\mathrm{reg}} \, \mathcal{L}_{\mathrm{reg}}
+ \lambda_{\mathrm{up}} \, \mathcal{L}_{\mathrm{up}}
+ \lambda_{\mathrm{smo}} \, \mathcal{L}_{\mathrm{smo}},
\end{aligned}
\end{equation}
where $\mathcal{L}_{\mathrm{gmof}}$ is the Geman--McClure robust loss~\cite{robustloss}.
The set $K$ contains 2D Sapiens keypoints, and $\hat{K}$ are the reprojected keypoints from SMPL-X.
In this stage we use all head, body, and hand keypoints whose confidence is larger than $0.6$.
Empirically, we observe that when the person stands with arms down, with hands invisible but forearms still visible, the predicted wrist locations are unstable and often flip upwards, which misleads the optimization.
Therefore, for upper-body inputs we ignore the left and right wrist keypoints when optimizing the body.
$M$ is the foreground mask obtained during preprocessing, and $\hat{M}$ is the rendered silhouette of the SMPL-X mesh; $\mathcal{L}_{\mathrm{mask}}$ encourages $\hat{M}$ to lie inside $M$.
$\mathcal{L}_{\mathrm{reg}}$ regularizes the current pose to stay close to the initial estimate.
For upper-body inputs, we set $\lambda_{\mathrm{up}} > 0$ and define $\mathcal{L}_{\mathrm{up}}$ to encourage the direction vector from the pelvis to the neck to be aligned with the vertical axis, correcting the front-leaning or backward-leaning poses caused by depth ambiguity in OSX.
When optimizing multiple consecutive frames jointly, $\mathcal{L}_{\mathrm{smo}}$ is defined as the second-order temporal difference of the projected 2D mesh vertices to reduce jitter.

\paragraph{FLAME Refinement.}
For video frames or single images, we integrate the GAGAvatar~\cite{chu2024gagavatar} tracking pipeline to obtain FLAME estimates.
Note that all frames in a video share the same FLAME shape $\boldsymbol{\beta}^{\text{flame}}$ and the same SMPL-X shape $\boldsymbol{\beta}^{\text{smplx}}$.
After obtaining FLAME predictions (shape $\boldsymbol{\beta}^{\text{flame}}$, expression $\boldsymbol{\psi}^{\text{flame}}$, and pose $\boldsymbol{\theta}^{\text{flame}}$), we estimate an affine transformation that aligns the canonical SMPL-X head to the predicted FLAME head, then replace the SMPL-X head vertices with the FLAME head before performing linear blend skinning.
If the FLAME tracking fails for a frame (\textit{e.g.}, facial landmark detection failed), we fall back to using a zero FLAME parameter as a dummy input.

We further refine the head by leveraging the dense FLAME re-projection like GUAVA tracker:
\begin{equation}
\begin{aligned}
\mathcal{L}^{\mathrm{head}} = {} &
\lambda_{\mathrm{head}} \, \mathcal{L}_{\mathrm{gmof}}\!\left(V_{\mathrm{head}}, \hat{V}_{\mathrm{head}}\right) \\
& + \lambda_{\mathrm{reproj}} \, \mathcal{L}_{\mathrm{gmof}}\!\left(K, \hat{K}\right) \\
& + \lambda_{\mathrm{mask}} \, \mathcal{L}_{\mathrm{mask}}\!\left(M, \hat{M}\right)
+ \lambda_{\mathrm{reg}} \, \mathcal{L}_{\mathrm{reg}} \\
& + \lambda_{\mathrm{up}} \, \mathcal{L}_{\mathrm{up}}
+ \lambda_{\mathrm{smo}} \, \mathcal{L}_{\mathrm{smo}},
\end{aligned}
\end{equation}
where $V_{\mathrm{head}}$ and $\hat{V}_{\mathrm{head}}$ denote the dense head vertices from FLAME and SMPL-X in image space, respectively.
In this stage we optimize the SMPL-X pose and translation
($\boldsymbol{\theta}^{\text{glob}}$, $\boldsymbol{\theta}^{\text{body}}$,
$\boldsymbol{\theta}^{\text{lhand}}$, $\boldsymbol{\theta}^{\text{rhand}}$, $\mathbf{t}$)
as well as the FLAME shape and expression
($\boldsymbol{\beta}^{\text{flame}}$, $\boldsymbol{\psi}^{\text{flame}}$).
The keypoint loss only supervises body keypoints in this stage.

\paragraph{Hand Refinement.}
Finally, we refine the hand regions.
If reliable hand observations are available, we run HaMeR~\cite{HaMeR} to estimate MANO parameters, from which we obtain dense hand keypoints and hand poses.
When valid hand poses are available, we update the SMPL-X hand poses
$\boldsymbol{\theta}^{\text{lhand}}, \boldsymbol{\theta}^{\text{rhand}}$ accordingly and perform a dedicated hand refinement with the objective
\begin{equation}
\begin{aligned}
\mathcal{L}^{\mathrm{hand}} = {} &
\lambda_{\mathrm{hand}} \, \mathcal{L}_{\mathrm{gmof}}\!\left(V_{\mathrm{hand}}, \hat{V}_{\mathrm{hand}}\right) \\
& + \lambda_{\mathrm{head}} \, \mathcal{L}_{\mathrm{gmof}}\!\left(V_{\mathrm{head}}, \hat{V}_{\mathrm{head}}\right) \\
& + \lambda_{\mathrm{reproj}} \, \mathcal{L}_{\mathrm{gmof}}\!\left(K, \hat{K}\right)
+ \lambda_{\mathrm{mask}} \, \mathcal{L}_{\mathrm{mask}}\!\left(M, \hat{M}\right) \\
& + \lambda_{\mathrm{reg}} \, \mathcal{L}_{\mathrm{reg}}
+ \lambda_{\mathrm{up}} \, \mathcal{L}_{\mathrm{up}}
+ \lambda_{\mathrm{smo}} \, \mathcal{L}_{\mathrm{smo}},
\end{aligned}
\end{equation}
where $V_{\mathrm{hand}}$ and $\hat{V}_{\mathrm{hand}}$ denote the dense hand vertices in image space.
In this stage we also optimize the SMPL-X poses of the wrists, shoulders, and elbows, while the keypoint loss supervises body and hand keypoints.

All three optimization stages use the Adam~\cite{adam} optimizer with a learning rate of $10^{-3}$. For additional implementation details, please refer to our released code.

\begin{table}[t]
  \centering
  \caption{Hyperparameters used for the three tracking stages.}
  \label{tab:tracking-hyperparams}
  \begin{tabular}{lc}
    \toprule
    Hyperparameter & Value \\
    \midrule
    $\lambda_{\mathrm{reproj}}\!\left(\mathcal{L}^{\mathrm{body}}\right)$ & $10^{2}$ \\
    $\lambda_{\mathrm{reproj}}\!\left(\mathcal{L}^{\mathrm{head}}\right)$ & $10^{2}$ \\
    $\lambda_{\mathrm{reproj}}\!\left(\mathcal{L}^{\mathrm{hand}}\right)$ & $10^{1}$ \\
    $\lambda_{\mathrm{reg}}\!\left(\mathcal{L}^{\mathrm{body}}\right)$    & $10^{2}$ \\
    $\lambda_{\mathrm{reg}}\!\left(\mathcal{L}^{\mathrm{head}}\right)$    & $10^{2}$ \\
    $\lambda_{\mathrm{reg}}\!\left(\mathcal{L}^{\mathrm{hand}}\right)$    & $10^{1}$ \\
    $\lambda_{\mathrm{mask}}$                                            & $10^{2}$ \\
    $\lambda_{\mathrm{up}}$                                              & $10^{4}$ \\
    $\lambda_{\mathrm{smo}}\!\left(\mathcal{L}^{\mathrm{body}}\right)$   & $5 \times 10^{2}$ \\
    $\lambda_{\mathrm{smo}}\!\left(\mathcal{L}^{\mathrm{head}}\right)$   & $5 \times 10^{4}$ \\
    $\lambda_{\mathrm{smo}}\!\left(\mathcal{L}^{\mathrm{hand}}\right)$   & $5 \times 10^{5}$ \\
    $\lambda_{\mathrm{head}}$                                            & $10^{3}$ \\
    $\lambda_{\mathrm{hand}}$                                            & $10^{2}$ \\
    \bottomrule
  \end{tabular}
\end{table}

\paragraph{Non-frontal Case.}
Estimating SMPL-X parameters for side and back views generated by Qwen-Image-Edit is particularly challenging, due to the lack of suitable pretrained models and supervision signals in these viewpoints.
Fortunately, these Qwen-generated side and back views are typically pose-neutral and almost perfectly aligned to $\pm 90^\circ$ and $180^\circ$ viewpoints.
We therefore introduce several tailored modifications when tracking Qwen-edited images.

For side views, we enforce the direction vectors between left and right ears, left and right shoulders, and left and right hips to be parallel to the camera viewing direction.
We disable the additional FLAME refinement and instead directly use Sapiens facial keypoints for optimization, because we observe that the GAGAvatar tracking pipeline tends to produce a noticeable tilt for $\pm 90^\circ$ side views.
Moreover, during the first body optimization stage, we do not optimize the global translation.
If translation is updated in this stage, the dense facial landmarks tend to pull the entire pose towards cases with a protruding neck.

For back views, reliable landmarks are largely unavailable.
In this case, we mainly rely on the silhouette loss $\mathcal{L}_{\mathrm{mask}}$ and the upright prior $\mathcal{L}_{\mathrm{up}}$ to obtain a plausible SMPL-X configuration.

\begin{figure*}[]
    \centering
    \includegraphics[width=0.75\textwidth]{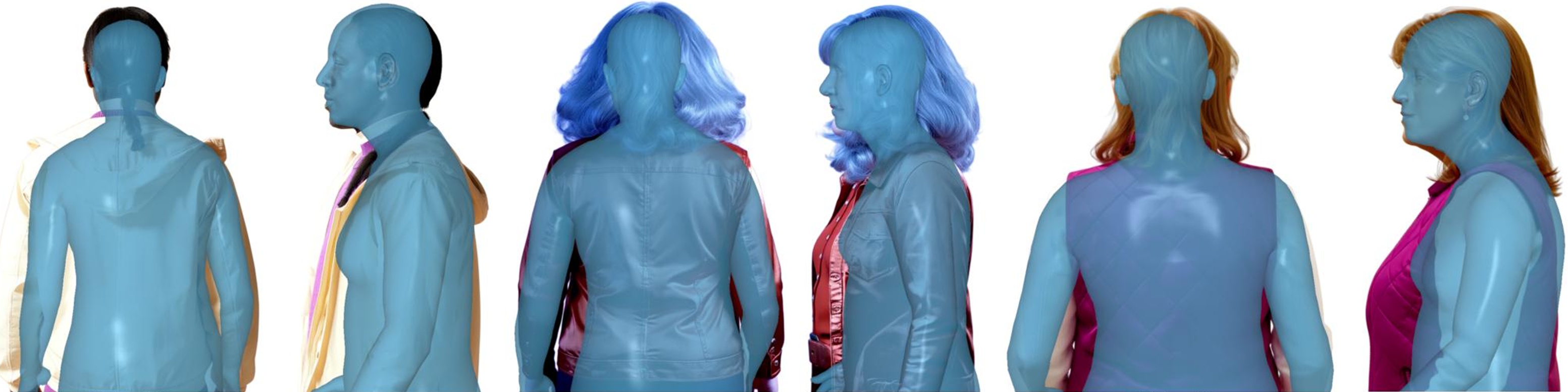}
    \caption{\textbf{Proxy Mesh Estimation}.
    Tracking results under side- and back-view poses.
    We visualize representative tracking outputs under challenging viewpoints.}
    \label{rebuttal: track}
\end{figure*}

We further visualize tracking results under side- and back-view poses in Fig.~\ref{rebuttal: track},
demonstrating the robustness of the tracking module under more challenging viewpoints.

\begin{figure*}[]
    \centering
    \includegraphics[width=0.75\textwidth]{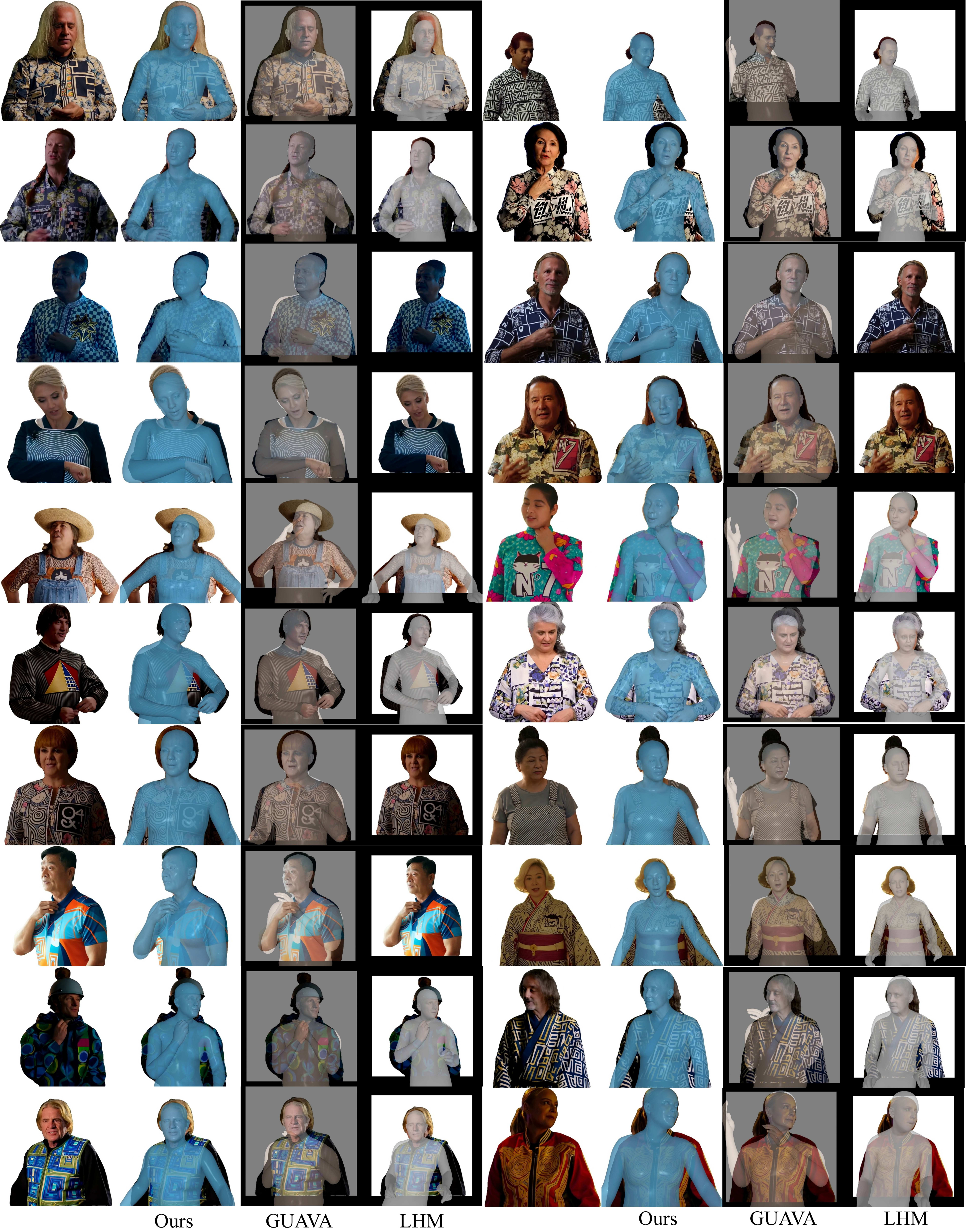}
    \caption{\textbf{Proxy Mesh Estimation}.
    We showcase how our tracker, GUAVA, and LHM perform on arbitrary upper-body images, highlighting the robustness under unconstrained input conditions.
    }
    \label{fig: supp_track}
\end{figure*}

\paragraph{Summary}
Our tracker is designed for the general case: it aims to provide stable estimates under diverse input conditions, thereby enabling us to scale up our dataset reliably.
We present qualitative comparisons with the LHM and GUAVA trackers in Fig.~\ref{fig: supp_track}. The LHM tracker fails to produce correct results in certain cases, while the GUAVA tracker can estimate accurately when both hands are clearly visible. In contrast, our tracker delivers robust performance across a wide range of input conditions.
\section{More Experiments}
\label{sec: supp_exp}

\subsection{Head Reenactment}
\label{sec:supp_head_exp}

\begin{figure*}[]
    \centering
    \includegraphics[width=0.99\textwidth]{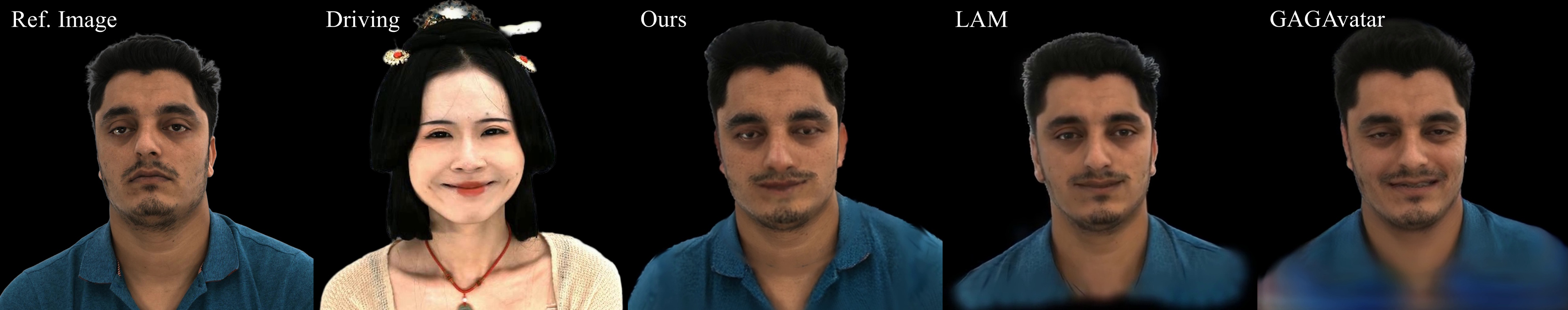}
    \caption{\textbf{Head Cross-Reenactment Results}.
    We show cross-reenactment examples in the head setting. Compared with other methods, our approach better preserves identity while maintaining stable motion transfer.}
    \label{rebuttal: head_cross_reenact}
\end{figure*}

We first provide additional quantitative comparisons on head reenactment.
Following prior works, we evaluate both self-reenactment and cross-reenactment.
For evaluation, we select 50 identities from RenderMe360~\cite{renderme360} and randomly sample one clip for each identity.
For cross-reenactment, we further randomly select 10 additional identities and sample one random clip from each of them as the driving sequences.
For self-reenactment, we report PSNR, SSIM, LPIPS, CSIM, average expression distance (AED), and average pose distance (APD).
For cross-reenactment, we report CSIM, AED, and APD.
We compare our method with LAM and GAGAvatar~\cite{chu2024gagavatar}.
Quantitative results are reported in Tab.~\ref{rebuttal: head_reenact}, and qualitative results of cross-reenactment are shown in Fig.~\ref{rebuttal: head_cross_reenact}.
It is worth noting that our pipeline is not specifically designed for head avatars, as the model is trained using only half-body data.
Nevertheless, it remains competitive in head reenactment and achieves favorable performance under cross-reenactment. Overall, the results suggest that our unified portrait animation framework transfers well to head reenactment scenarios,
despite not being trained under a head-only setting.

\subsection{Failure Cases}
\label{sec:supp_failure_cases}

\begin{figure*}[]
    \centering
    \vspace{-4mm}
    \includegraphics[width=0.99\textwidth]{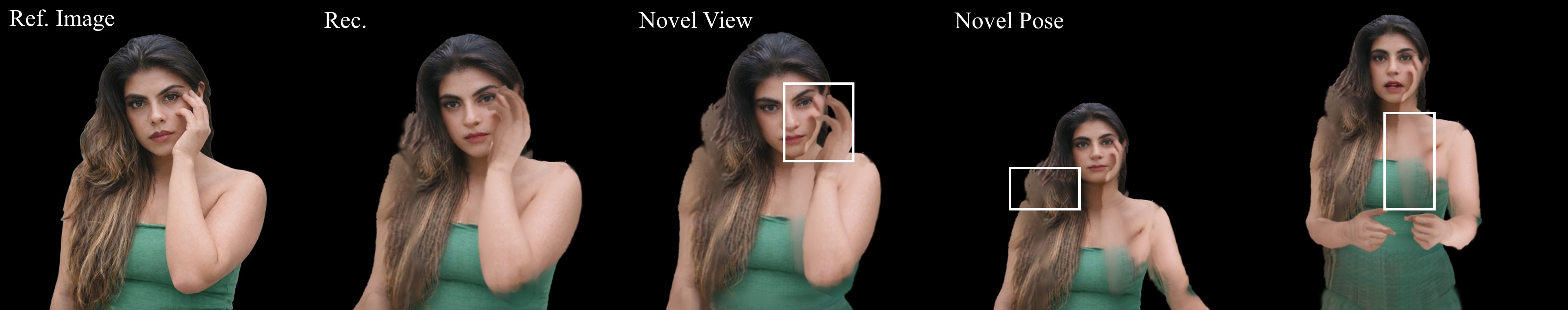}
    \caption{\textbf{Failure Cases}.
    We highlight challenging scenarios such as heavy occlusion and long or complex hair.}
    \label{rebuttal: failure_case}
\end{figure*}

We show representative failure cases in Fig.~\ref{rebuttal: failure_case}.
Typical failure modes include heavy occlusion and long or complex hair.
These cases remain challenging because the observable evidence is incomplete and the geometry/appearance ambiguity becomes significantly larger.
Addressing these scenarios may require stronger priors, more diverse training data, and improved temporal or geometric constraints.

\subsection{Dynamic Pose and Novel-View Synthesis}
\label{sec:supp_dynamic_pose_novel_view}

\begin{figure*}[]
    \centering
    \includegraphics[width=0.99\textwidth]{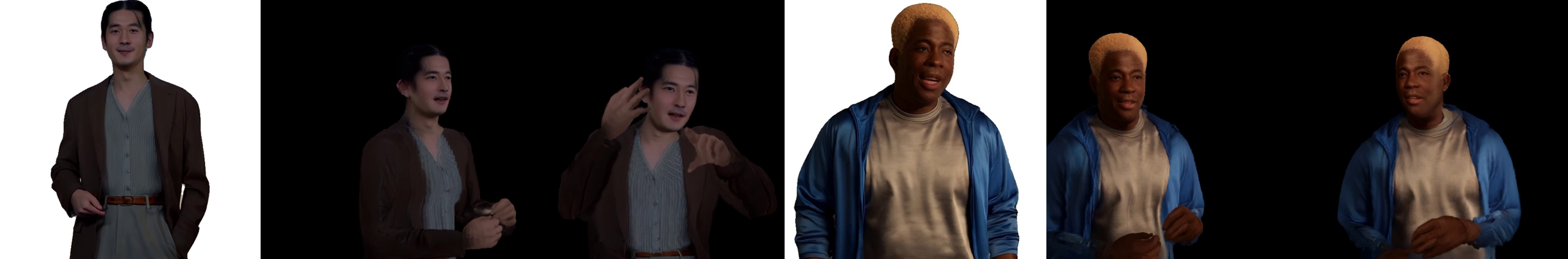}
    \caption{\textbf{Dynamic-pose animation and novel-view synthesis results}.
    We show representative examples under challenging pose changes and across a wide range of viewpoints.
    }
    \label{rebuttal: dyn_novel_view}
\end{figure*}

We provide additional results on dynamic-pose animation
and extensive novel-view synthesis.
As shown in Fig.~\ref{rebuttal: dyn_novel_view}, our method remains stable under pose variation and
produces plausible renderings across a wide range of viewpoints.

\subsection{Qualitative Ablation}
\label{sec:supp_qualitative_ablation}

We additionally provide qualitative ablations on the proposed design choices, including the use of Shell-UV and the depth of the decoder.
In practice, minor residual tracking misalignment may reduce the sensitivity of standard image-space metrics,
which can lead to relatively small numerical differences across ablations.
Therefore, qualitative comparisons are particularly informative for revealing perceptual improvements.
As shown in Fig.~\ref{rebuttal: ablation}, the full model produces the most faithful appearance and the richest local details.

\subsection{Full-Body Animation}
\label{sec:supp_full_body}

\begin{figure*}[]
    \centering
    \includegraphics[width=0.99\textwidth]{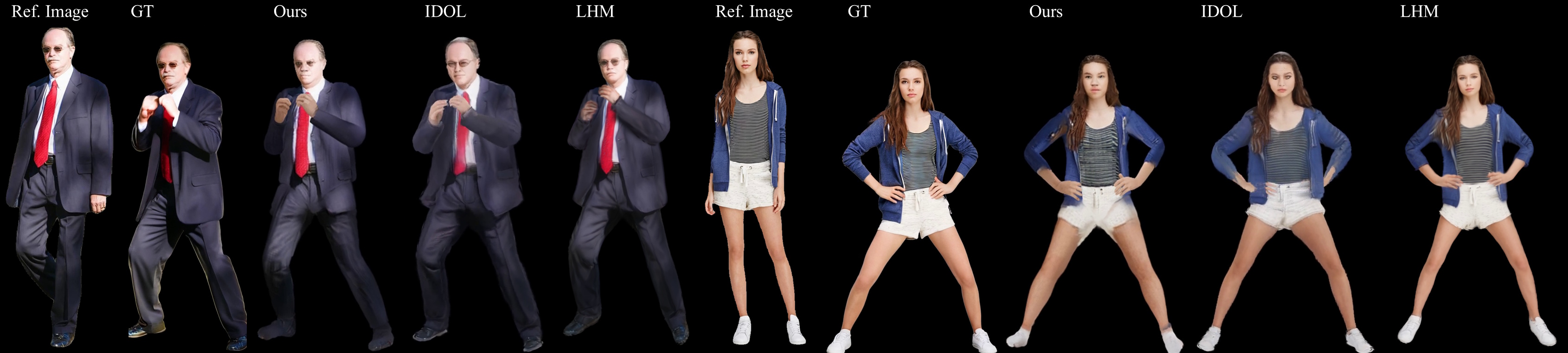}
    \caption{\textbf{Full-body animation results}.
    Although our model is trained using only upper-body data, it generalizes to full-body animation and produces temporally coherent motion with plausible appearance synthesis.
    }
    \label{rebuttal: full_body_animation}
\end{figure*}

Although the inference checkpoint used in these experiments is trained only on upper-body data, our pipeline naturally supports
head-only, upper-body, and full-body inputs, and can generalize to full-body animation scenarios.
Figure~\ref{rebuttal: full_body_animation} presents additional full-body animation results.

\begin{table*}[]
\caption{
\centering
\textbf{Results on Head Reenactment}. We show both self-reenactment and cross-reenactment results together with comparisons to baseline methods.
}
\label{rebuttal: head_reenact}
\centering
\scalebox{0.99}{
\begin{tabular}{c|cccccc|ccl}
\hline
                      & \multicolumn{6}{c|}{Self Reenactment}                                                                                                                                 & \multicolumn{3}{c}{Cross Reenactment}                       \\
\multicolumn{1}{l|}{} & \multicolumn{1}{l}{PSNR↑} & \multicolumn{1}{l}{SSIM↑} & \multicolumn{1}{l}{LPIPS↓} & \multicolumn{1}{l}{CSIM↑} & \multicolumn{1}{l}{AED↓} & \multicolumn{1}{l|}{APD↓} & \multicolumn{1}{l}{CSIM↑} & \multicolumn{1}{l}{AED↓} & APD↓ \\ \hline
Ours                  & \cb{19.04}                          & \cb{0.8526}                          & \cb{0.1613}                           & \clb{0.7029}                          & \clb{0.1319}                         & \cb{0.0499}                          &  0.6161                         & \cb{0.2720}                        & \cb{0.1299}     \\
LAM                   & 17.19                          & 0.7526                          & 0.2207                           & 0.6994                          & 0.1565                         & 0.1080                          & \clb{0.6248}                          & 0.2839                         & 0.1464     \\
GAGAvatar             & \clb{18.48}                          & \clb{0.7877}                          & \clb{0.1872}                           & \cb{0.7294}                          & \cb{0.0964}                         & \clb{0.0698}                          & \cb{0.6536}                          & \clb{0.2782}                         & \clb{0.1455}     \\ \hline
\end{tabular}
}
\end{table*}

\begin{figure*}[]
    \centering
    \includegraphics[width=\textwidth]{fig/supp_ablation.jpg}
    \caption{\textbf{Qualitative Ablation}.
    We visualize the impact of Shell-UV and decoder depth. The full model yields the most faithful appearance and detail.
    }
    \label{rebuttal: ablation}
\end{figure*}


\end{document}